\documentclass[runningheads]{llncs}
\usepackage{graphicx}
\usepackage{caption}
\usepackage{subcaption}
\graphicspath{figure}
\usepackage{amsmath,amssymb} 
\usepackage{authblk}
\usepackage{color}
\usepackage[width=122mm,left=12mm,paperwidth=146mm,height=193mm,top=12mm,paperheight=217mm]{geometry}
\begin{document}

\author[1]{Sen He}
\author[2]{Ali Borji}
\author[1]{Yang Mi}
\author[1]{Nicolas Pugeault}

\title{What Catches the Eye? Visualizing and Understanding Deep Saliency Models} 



\author{Sen He \textsuperscript{1}, Ali Borji \textsuperscript{2}, Yang Mi \textsuperscript{1}, Nicolas Pugeault \textsuperscript{1}}
\institute{\textsuperscript{1} Department of Computer Science\\University of Exeter\\ %
\textsuperscript{2} Center for Research in Computer Vision\\University of Central Florida \\ %
}
\maketitle

\begin{abstract}
Deep convolutional neural networks have demonstrated high performances for fixation prediction in recent years. How they achieve this, however, is less explored and they remain to be black box models. Here, we attempt to shed light on the internal structure of deep saliency models and study what features they extract for fixation prediction. Specifically, we use a simple yet powerful architecture, consisting of only one CNN and a single resolution input, combined with a new loss function for pixel-wise fixation prediction during free viewing of natural scenes. We show that our simple method is on par or better than state-of-the-art complicated saliency models. Furthermore, we propose a method, related to saliency model evaluation metrics, to visualize deep models for fixation prediction. Our method reveals the inner representations of deep models for fixation prediction and provides evidence that saliency, as experienced by humans, is likely to involve high-level semantic knowledge in addition to low-level perceptual cues. 
Our results can be useful to measure the gap between current saliency models and the human inter-observer model and to build new models to close this gap.

\end{abstract}

\newcommand{\citet}[1]{\citeauthor{#1} (\citeyear{#1})}

\section{Introduction}
The human visual system receives a large amount of information every second (about $10^{8}$ to $10^{9}$ bits). An essential mechanism that allows the human visual system to process such a vast amount of information in real time is its capacity to selectively focus attention on parts of the scene. This process has been extensively studied by Psychologists to discover which visual patterns capture human attention. Desimone \& Duncan \cite{desimone1995neural} found that parts of an image that differ from their surroundings stand out. This paradigm is called \textit{center-surround difference} in early computational modeling of visual attention. Based on the center-surround difference and the feature integration theory proposed by Treisman \& Gelade \cite{treisman1980feature}, many computational models of visual attention have been proposed \cite{itti2000saliency,harel2007graph,garcia2012saliency}. 

In recent years, with the availability of large scale datasets recording mouse movements of human subjects as a proxy of gaze (e.g., \cite{jiang2015salicon}) and of powerful parallel hardware, the development of data driven approaches based on deep learning have demonstrated significantly higher performance than previous models on all benchmarks \cite{mit-saliency-benchmark}. Currently, almost all deep saliency models treat the gaze map as a small scale map recording the density of fixations at every image location (downsampled from the ground truth \cite{huang2015salicon,jetley2016end}). Such models are almost invariably trained by minimizing the distance between the predicted saliency maps and the ground truth. At inference time, the saliency map is then upsampled to the input's image size. Such deep saliency models have achieved much better performance than  models based on hand-crafted features or psychological assumptions, but unlike for the task of image recognition, where the representations learned by deep neurons have been studied and visualized~\cite{zeiler2014visualizing,bau2017network}, it remains unclear why deep saliency models perform so well or what salient patterns have deep neurons attuned to in the process. The complexity of some of the proposed architectures make them even more inscrutable. 

In this paper, we use a simple yet powerful residual-like decoder with a new loss function for pixel-wise gaze prediction. The architecture is similar to the architecture in \cite{pan2017salgan}, but we dispense with the GAN training and instead propose a simpler, residual decoder. We demonstrate that the model although simpler, achieves better performance on most metrics and datasets. Additionally, we propose a novel method to visualize  and analyze the representations learnt by deep saliency models. To the best of our knowledge, this is the first work which looks inside deep saliency models. 

The rest of the paper is organized as follows. Section 2 reviews the state-of-the-art gaze prediction models as well as visualization methods for deep convolutional networks. Section 3 introduces the proposed deep saliency model and the model visualization method. Experimental results and benchmarks are presented in Sections 4 and 5. 

\section{Related work}
In this section, we first review the state-of-the-art deep gaze prediction models before introducing  visualization methods for deep convolutional networks.

\subsection{Deep saliency models}
The release of SALICON dataset \cite{jiang2015salicon}, offered for the first time a large scale dataset for saliency, which spurred the development of a number of saliency models. 
For example, Deepnet \cite{pan2016shallow} learns saliency using 8 convolutional layers, where only the first 3 layers were initialized from the pre-trained image classification model. 
PDP \cite{jetley2016end} treats the gaze map as a small scale probability map. Authors investigated different loss functions for training their gaze prediction model and found that Bhattacharyya distance is the best loss function when the gaze map is treated as a small scale probability map. 
The Salicon \cite{huang2015salicon} model uses multi resolution inputs, and combines feature representations in the deep layers for gaze prediction. 
Deepfix \cite{kruthiventi2017deepfix} combined the deep architectures of VGG \cite{simonyan2014very}, Googlenet \cite{szegedy2015going}, and Dilated convolutions \cite{yu2015multi} in their network as well as adding a central bias, to achieve a higher performance than previous models. SalGAN \cite{pan2017salgan} uses an encoder-decoder architecture and proposes the binary cross entropy (BCE) loss function to perform pixel-wise (rather than image-wise) saliency estimation. After pre-training the encoder-decoder, they use a Generative Adversarial Network (GAN) to boost their model's performance. 
DVA \cite{wang2017deep} uses multiple deep layer representations, builds a decoder for each layer, and fuses them at the final stage for pixel-wise gaze prediction. 
DSCLRCN \cite{liu2016deep} also uses multiple inputs by adding a contextual information stream, and concatenates the original representation and the contextual representation into a LSTM network for the final prediction. 

Table~\ref{tab:modelm} provides a comparison of state-of-the-art deep saliency models. 
Complex architectures \cite{huang2015salicon,kruthiventi2017deepfix,wang2017deep,cornia2016predicting,liu2016deep} are intrinsically inscrutable and difficult to interpret, hence in this article we propose to use a simple fully convolutional encoder with a residual decoder, using the exponential absolute distance (EAD) to do pixel-wise gaze prediction. We demonstrate that despite its simplicity, this architecture can compete, or even outperform, more complex state-of-the-art architectures. 

\begin{table}[h!]
\centering
\begin{tabular}{|c|c|c|c|c|c|c|}
\hline

Model                       &Input                      &CNN                                  &LSTM   &CB        &Loss & pixel/PD\\
\hline
\hline
DSCLRCN~\cite{liu2016deep}              &multi inputs            &Resnet \cite{he2016deep}, Places \cite{zhou2014learning}                    &Yes      &no       &NSS&pixel\\
Deepfix~\cite{kruthiventi2017deepfix}                   &single input     &MA(VGG,Googlenet,Dilated) &no  &yes  &$L^2$&pixel\\
Salicon~\cite{huang2015salicon}                  &MR inputs               &VGG                                   &no  &no    &K-L & PD\\
SalGAN~\cite{pan2017salgan}            &single input                  &VGG,GAN                         &no    &no     &BCE&pixel\\
PDP~\cite{jetley2016end}                  &single input                 &VGG                               &no        &no     &Bha&PD\\
DVA~\cite{wang2017deep}                  &single input                  &VGG,MD                        &no       &no       &BCE&pixel\\
Deepnet~\cite{pan2016shallow}           &single input                  &Custom 8-layers                                &no       &no       &$L^2$&pixel\\
\hline
Ours                  &single input                 &VGG                           &no           &no       &EAD&pixel\\
\hline
\end{tabular}
\caption{\small{
Comparison of saliency prediction models. MA: multi-architecture, MR: multi-resolution, PD: probability distribution based, Bha: Bhattacharyya distance, MD: multiple decoders, CB: central bias, K-L: Kullback-Leibler divergence.
}
}
\label{tab:modelm}
\end{table}

\vspace{-1.5cm}

\subsection{Visualizing deep neural networks}
The success of deep convolutional neural networks has raised the question of what representations are learned by neurons located in deep layers. One approach towards understand how CNNs work and learn is to visualize individual neurons' activations and receptive fields. For example, Zeiler \& Fergus~\cite{zeiler2014visualizing} proposed a deconvolution network in order to visualize the original patterns that activate the corresponding activation maps. In the forward pass of a convolutional neural network the main operations are convolution, ReLU (or another nonlinearity) and pooling. Conversely, a deconvolution network is consists of the three steps of unpooling, transposed convolution (using the pre-trained weights in the forward pass, and transposing them for convolution), and the ReLU operation. 
Yosinski et al.~\cite{yosinski2015understanding} developed two tools for understanding the deep convolutional neural networks. The first of these tools is designed to visualize the activation maps at different layers for a given input image. The second tool aims to estimate the input pattern which a network is maximally attuned to for a given object class. In practice, the last layer of a classification deep neural network typically features one neuron per object class. Yosinski et al. propose to use gradient ascent (with regularization) to find the input image that maximizes the output of a specific neuron (i.e., for a specific object class). Hence, it derives the optimum input that appeals to the network for a specific class.

Both visualization methods discussed above are essentially qualitative. In contrast, Bau et al.~\cite{bau2017network} proposed a quantitative method to give each activation map a \textit{semantic meaning}. 
In their work, they proposed a dataset with 6 image categories and 63,305 images for network dissection, where
each image is labeled with pixel-wise semantic meaning. At first, they forward all images in the dataset into a pre-trained deep model. For each activation map inside the model, different inputs have different patterns. Then, they compute the distribution of each unit activation map over the whole dataset, and determine a threshold for each unit based on its activation distribution. With the threshold for each unit, the activation map for each input image is quantized into a binary map. Finally, they compute the intersection over union (IOU) between the quantized activation map and the labeled ground truth to determine what objects or object parts that unit is detecting. 

Although these approaches provide useful insight into the workings of deep neural networks, they are ill-suited for understanding deep saliency networks: If it is reasonable to expect that neurons in a dog/cat classifier will encode patterns characteristic of dogs and cats, a saliency model is expected to encode both salient patterns but \textit{also} non-salient ones.
For this reason, we propose to use the normalized scan-path saliency (NSS) \cite{bylinskii2016different} score to determine whether individual neurons act as negative or positive predictors of gaze in the network. Moreover, in order to interpret what has been learnt as salient by the model, we use the network dissection approach of Bau et al.~\cite{bau2017network} to highlight what objects or object parts neurons in our saliency models are implicitly attuned to. 

\section{Methodology}
This section will first introduce the proposed simplified architecture for saliency estimation. In a second part, we then describe how to visualize and analyze deep saliency models.

\subsection{Gaze prediction}\label{sec:model}
\begin{figure}[h!]
\vspace*{-30pt}
\centering
\includegraphics[width=10cm]{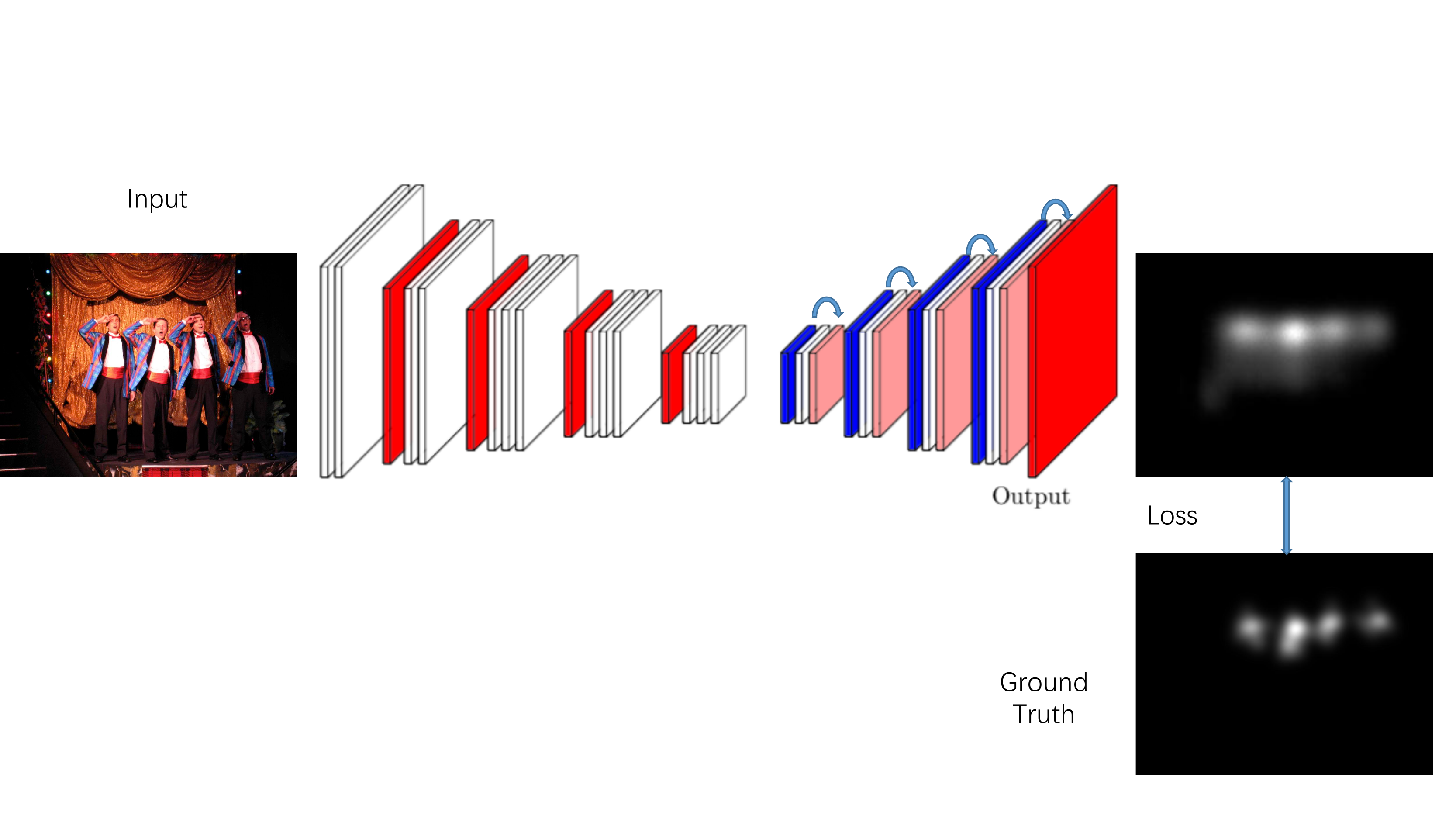}
\caption{The encoder and residual-decoder architecture of our network.}
\label{fig:model}
\end{figure}

The whole architecture of our network is illustrated in Figure~\ref{fig:model}. The input is first processed 
by encoder network, and represented by a feature tensor ($F^0$) of shape $M\times N\times K$.
\begin{align}
F^0 = CNN (Image) = \left[ X_{1},X_{2},\cdot \cdot \cdot,X_{M\times N}\right],
\end{align}
where $M\times N$ is the number of locations in the feature tensor, and $K$ is the dimension of each location. In our model, we use the first 5 convolutional blocks (we removed the last pooling layer, and kept the first 4 pooling layers in the encoder) of a pre-trained VGG16 \cite{simonyan2014very} network to initialize the feature extraction part and fine-tune it during training. The input was resized to $240\times 320$, hence the shape of the feature tensor is $15\times 20\times 512$.

After feature extraction, the feature tensor was then fed into the residual-decoder. The decoder is consists of four blocks, where each block upsamples the feature tensor once to recover the resolution lost in the encoding stage. Each block shares three similar processes: convolution for dimension reduction, normal convolution, and deconvolution to recover the resolution lost in the encoder due to pooling.

In each block, the feature tensor from the previous block $F^{n-1}$ is first processed by a dimension reduction convolutional layer $C^n_1$ to reduce the number of feature maps $K$. In our model, we halve the number of feature maps $K$ in each block of the decoder. 
\begin{align}
F^{n}_1 = C^n_1 (F^{n-1}) 
\end{align}
Then, the processed feature tensor ($F^n_1$) is processed by a conventional convolutional layer $C^n_2$ for further processing.
\begin{align}
F^n_2 = C^n_2(F^n_1) 
\end{align}
Finally, the two processed tensors ($F^n_1$ and $F^n_2$) are added together and then sent to a deconvolutional layer $D^n$ to increase the tensor resolution and generate the block's output tensor $F^n$.
\begin{align}
F^n = D^n (F^n_1+F^n_2)
\end{align}
The kernel size was set to $3\times 3$ for convolutional layers and $2\times 2$ for deconvolutional layers. Zero-padding was used to preserve the input's scale. The last layer of the decoder is a $3\times 3$ convolutional layer, which transforms the $64$ (output of last deconvolutional layer) activation maps into the saliency map. No further processing was implemented.

To train our model, we propose a new pixel-wise loss function, the exponential absolute distance (EAD), formulated as follows:
\begin{align}
EAD = \sum_{i=1}^N (\exp|p_{i}-g_{i}| -1)
\end{align}
where, $N$ is the number of pixels in the gaze map, $p_{i}$ and $g_{i}$ is the prediction and ground truth at the $i_{th}$ pixel.
\begin{figure*}[h!]
\vspace*{-15pt}
	\begin{center}
		\begin{subfigure}{0.32\textwidth}
			\includegraphics[width=\textwidth]{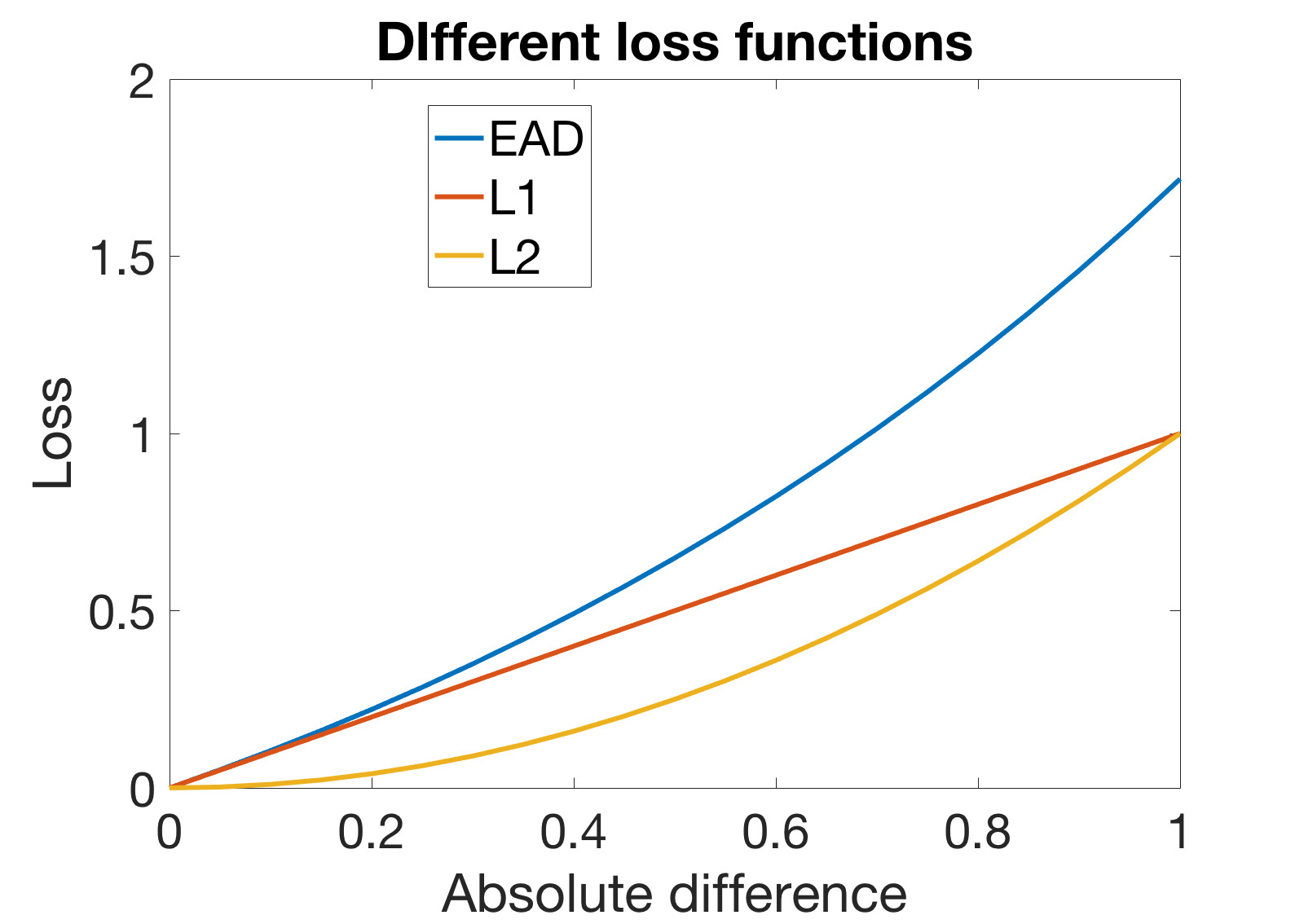}
			\caption{3 loss functions}
		\end{subfigure}
		\begin{subfigure}{0.32\textwidth}
			\includegraphics[width=\textwidth]{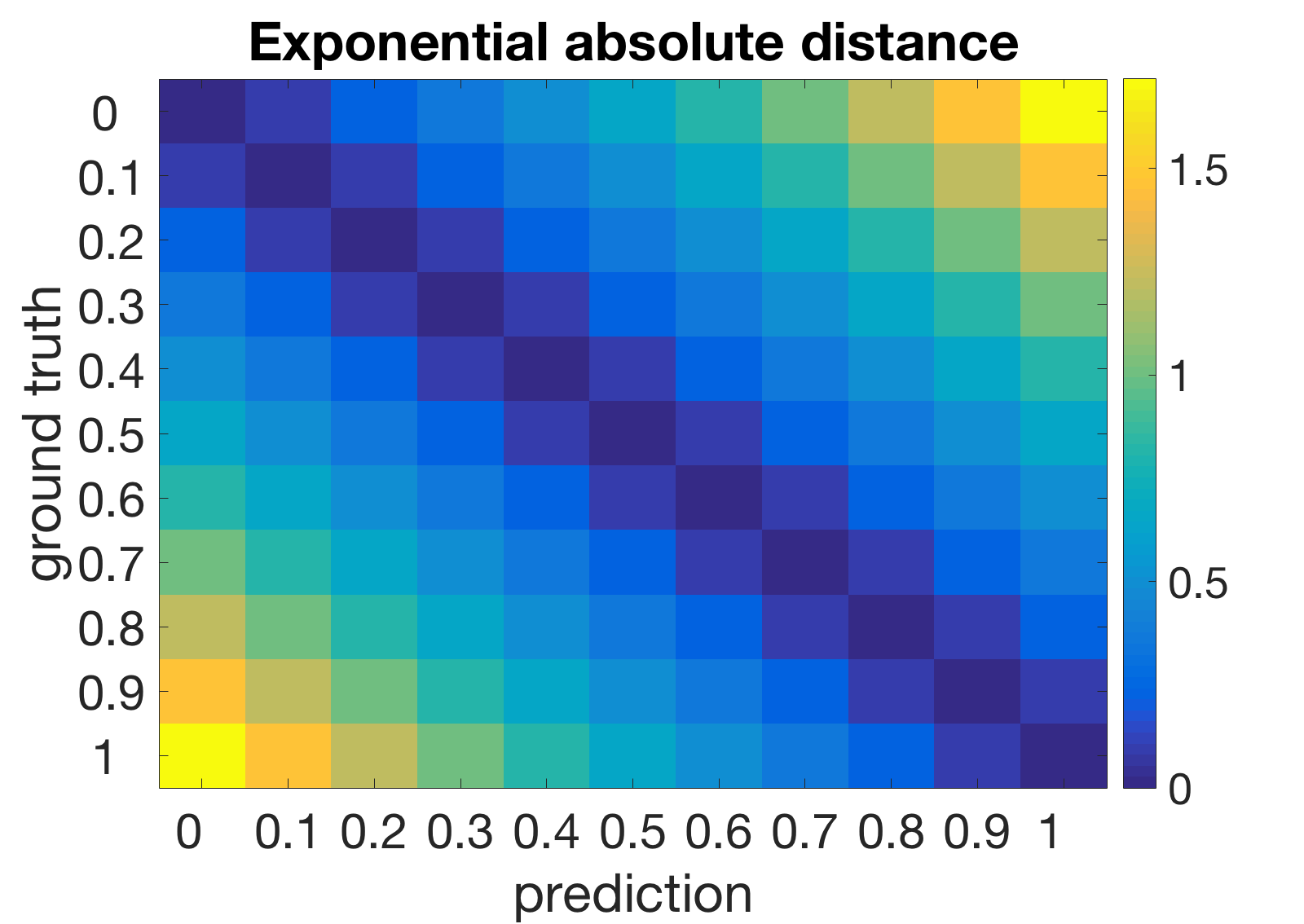}
			\caption{EAD loss map}
		\end{subfigure}
		\begin{subfigure}{0.32\textwidth}
			\includegraphics[width=\textwidth]{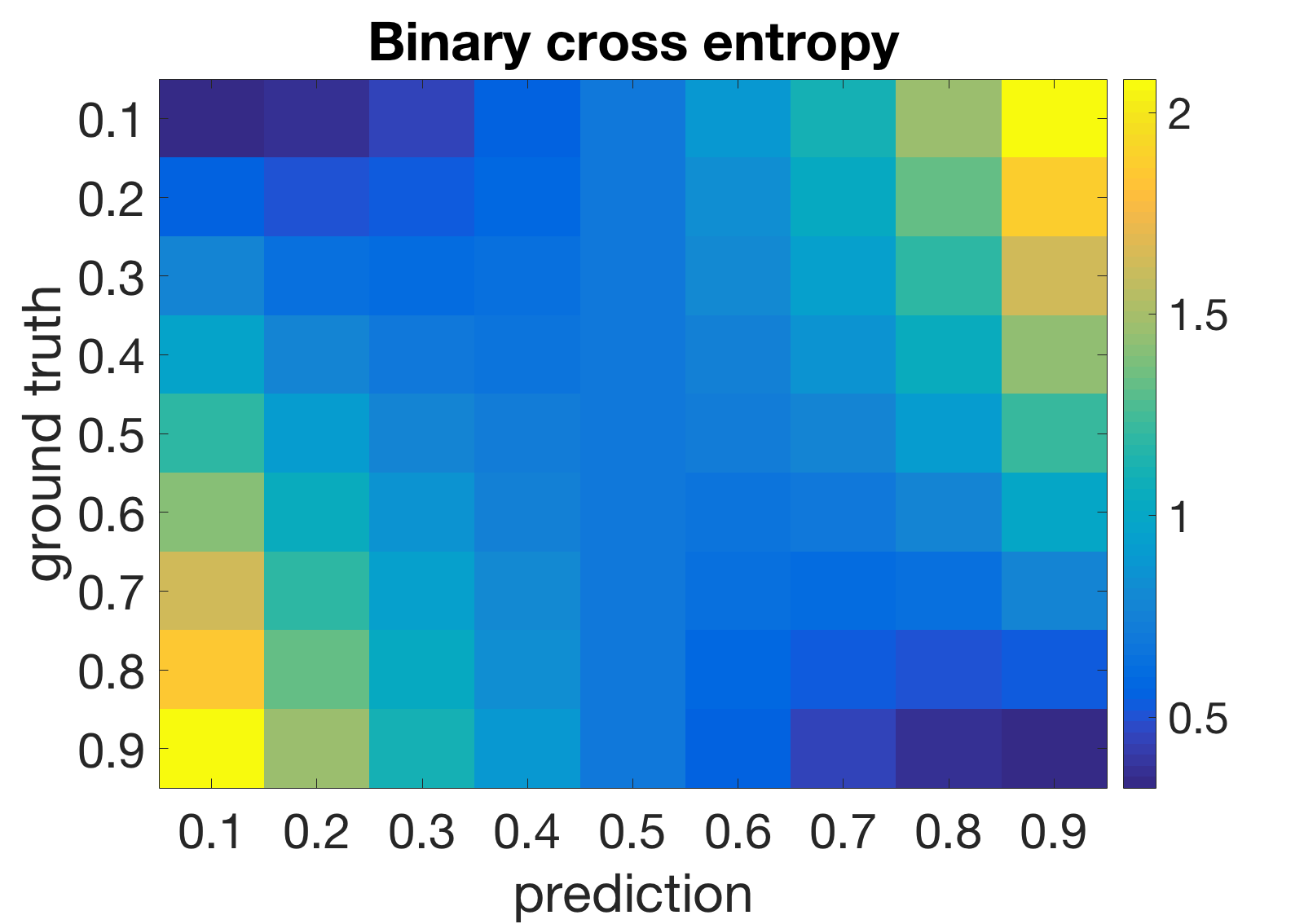}
			\caption{BCE loss map}
		\end{subfigure}
	\end{center}
	\caption{Properties of different loss functions.}
	\label{fig:loss_func}
\end{figure*}
Compared to the $L^2$ distance, the EAD has a better gradient when the absolute difference is small. Compared to the $L^1$ distance, which is linear in the absolute difference, the EAD gives a larger punishment when the difference is large. 
In contrast to EAD, the BCE loss proposed in \cite{pan2017salgan} yields a non-zero loss even for perfect predictions (as illustrated in Figure~\ref{fig:loss_func}). 
The unbounded nature of the BCE requires the application of an additional sigmoid function to produce pixel-wise saliency values in the range [0,1].
The model is trained using Tensorflow \cite{abadi2016tensorflow} with the Adam \cite{kingma2014adam} optimizer. We set the initial leaning rate as $5\times 10^{-5}$, and decay it with a factor of $0.1$ after each training epoch.

\subsection{Model visualization} \label{sec:viz}
As discussed before, one important question is what is learnt by a deep saliency model that allows it to outperform hand-crafted shallow models based on psychological theories? In other words, what specific salient patterns are learnt by the model? One hypothesis is that such deep network encode semantic information about saliency going beyond classical centre-surround assumptions. 

\begin{figure}[h!]
\centering
\includegraphics[width=10cm]{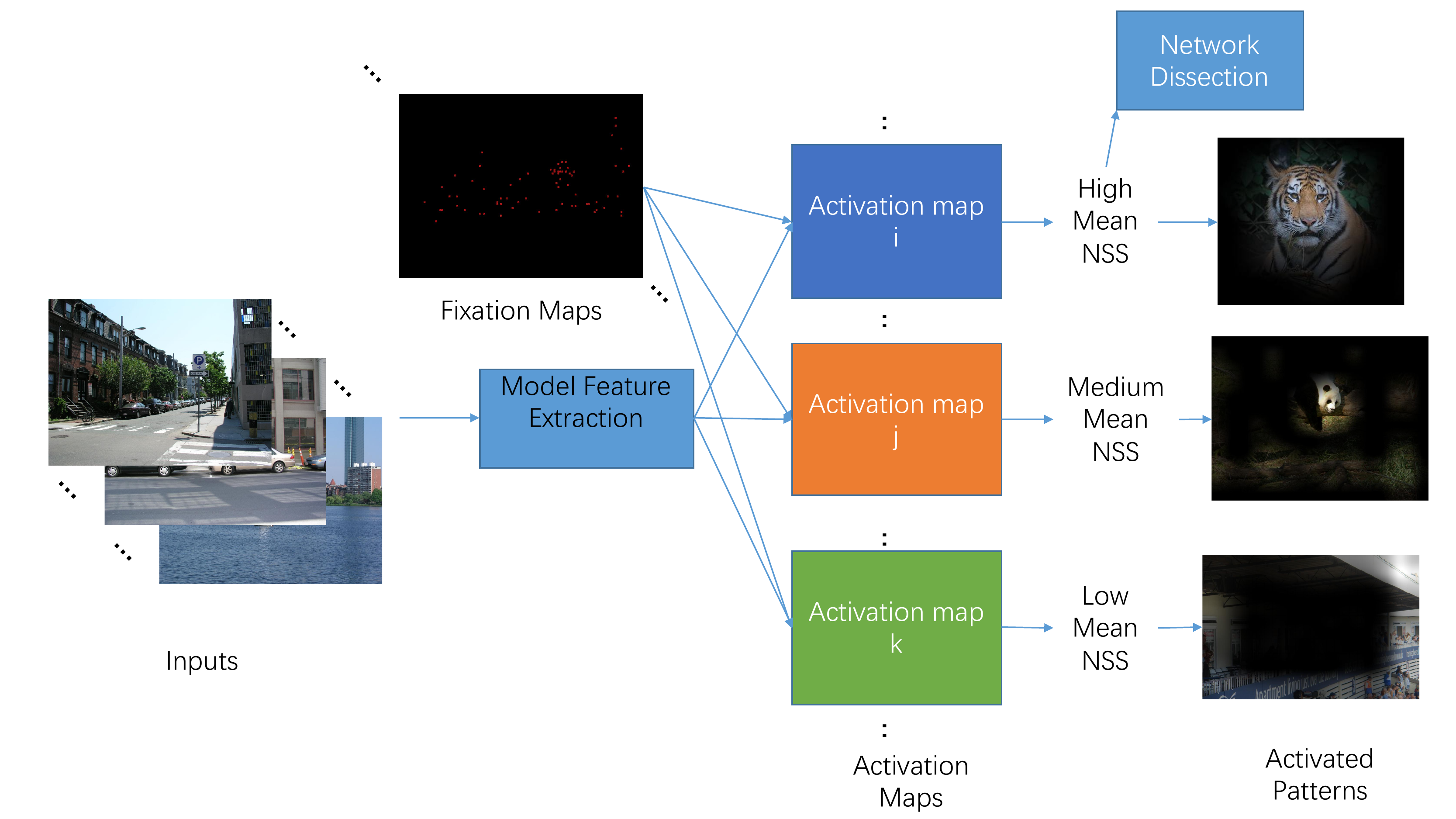}
\vspace*{-12pt}
\caption{The visualization method to compute the NSS score for each unit activation map.}
\label{fig:visualisation}
\end{figure}


Here, we propose to use the actual saliency evaluation metric, the normalized scan-path saliency (NSS) score, to visualize and understand inner representations of deep saliency models. At first, we feed all images in the dataset with  fixations (MIT1003 dataset \cite{judd2009learning}) to the pre-trained deep saliency model. For each single image, it produces a set of activation maps as the output of the feature extraction part (in our model, this is the output of the encoder), one per neuron. Each activation map has a unique pattern for a given input image. We rescale the activation map to the input's scale and use the activation to compute the NSS score for each neuron over the whole dataset. We use the top 5 NSS scores for each unit activation map, and compute their mean as the mean NSS score of each unit activation map (As a convolutional feature channel can only correspond to a certain type visual pattern \cite{simon2015neural,zhang2016picking}). Therefore, each neuron's activation map has a mean NSS score, which indicates its correlation with human gaze locations. Using the mean NSS score for each neuron's activation map, we normalize the mean NSS score across all activation maps between $[0,1]$, and set a threshold $T$ (we choose $T=0.9$ in our experiment). Neurons with mean NSS score above threshold $T$ are identified as \textit{positive fixation detectors}. 

After selecting positive fixation detectors, we use network dissection \cite{bau2017network} (using the same method and dataset as the authors) to reveal what kind of object or object part are those positive fixation detectors attuned to. We proceed as follows: For every image in the Broden dataset \cite{bau2017network}, there is a unique pattern for each unit activation map. For each neuron's activation map, we compute the distribution of its values over the whole dataset and find the threshold $T_k$ such that the value larger than $T_k$ with a probability $>0.005$. 
Then, all activation maps for all images are scaled to the input size and are quantized to a binary map. Finally, the IOU~\cite{bau2017network} is computed for each activation map to determine what sort of objects or object parts they are attuned to detecting (more details are in the \cite{bau2017network}). In our work, we only show the objects or object parts for the positive fixation detectors.

\section{Saliency prediction performance}
In this section, we first introduce the datasets used in the experiments and then show the performance of different pixel-wise loss functions as well as the comparison between our model and other state-of-the-art models.
\subsection{Datasets}
\begin{description}
\item{SALICON \cite{jiang2015salicon}:} The SALICON dataset is the largest dataset in the field of visual saliency. Saliency maps are estimated from  human observers' mouse clicks gathered over 20,000 images, with 10,000 images in the training set, 5,000 images in the validation dataset, and another 5,000 images in the testing dataset. We use the SALICON training and validation datasets to train and validate our model.

\item{MIT1003 \cite{judd2009learning}:} This dataset includes gaze data of 15 subjects using an eye tracker over 1,003 images. It is used in the visualization part. To compare our model with other state-of-the-art models on MIT300 benchmark, we also randomly choose 900 images from this dataset to fine-tune our model and another 103 images for testing the performance of different loss functions.

\item{MIT300 \cite{judd2012benchmark}:} This dataset is the standard benchmark dataset for human gaze prediction. It includes the gaze data of 39 subjects over 300 images.

\item{Broden \cite{bau2017network}:} This dataset contains 63,305 images. With four subsets, ADE20K (22,211 images), Opensurfaces (25,351 images), DTD (5,639 images), and PASCAL(10,104 images), with pixel-wise semantic labels. It is used for network dissection.
\end{description}

\subsection{Model performance}
Table~\ref{tab:loss} records the accuracy of predicted saliency maps according to a range of standard error measures: Normalised Scan-path (NSS), Cross Correlation (CC), Area Under ROC curve (AUC) and Similarity (Sim) (we refer to \cite{bylinskii2016different} for a discussion of saliency metrics). 
The accuracy is recorded for different pixel-wise loss functions. We can see that our proposed EAD loss achieves the best performance among all pixel-wise loss functions. Furthermore, the $L^2$ loss function, which is used as a baseline loss function in many deep gaze prediction models \cite{jetley2016end,pan2017salgan}, also shows good performance, which demonstrates that the proposed architecture is competitive regardless of the loss function and despite its comparative simplicity. This experiment was performed on the MIT1003 dataset.  

\begin{table}[h!]
\centering
\begin{tabular}{|c|c|c|c|c|}
\hline
Loss function             &NSS        &CC                       &AUC     &Sim\\
\hline
\hline
L1                              &2.388             &0.684              &0.855     &0.556\\
L2                              &2.389             &0.686              &\textbf{0.881}      &0.532\\
BCE                          &2.083             &0.614              &0.851      &0.488\\
\hline
EAD(proposed)                 &\textbf{2.404}             &\textbf{0.701}             &0.869         &\textbf{0.570}\\
\hline
\end{tabular}
\caption{The performance of different loss functions on MIT1003 testing dataset.}
\label{tab:loss} 
\end{table}

\begin{figure*}
	\begin{center}
		\begin{subfigure}{0.15\textwidth}
			\includegraphics[width=\textwidth]{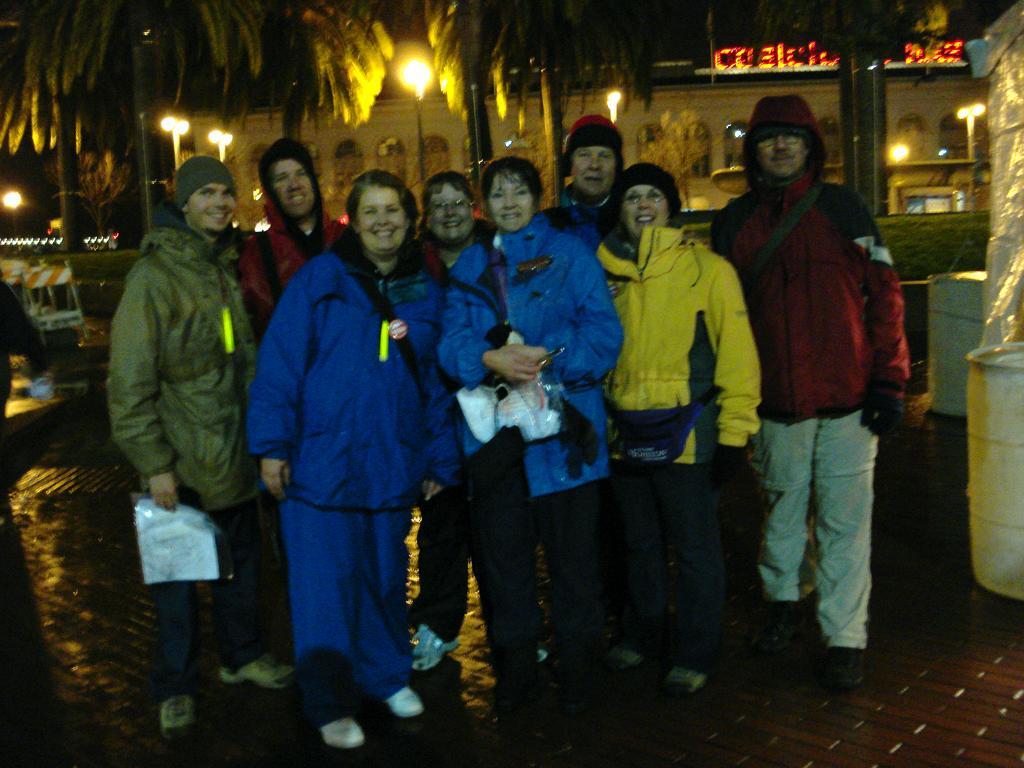}
		\end{subfigure}
		\begin{subfigure}{0.15\textwidth}
			\includegraphics[width=\textwidth]{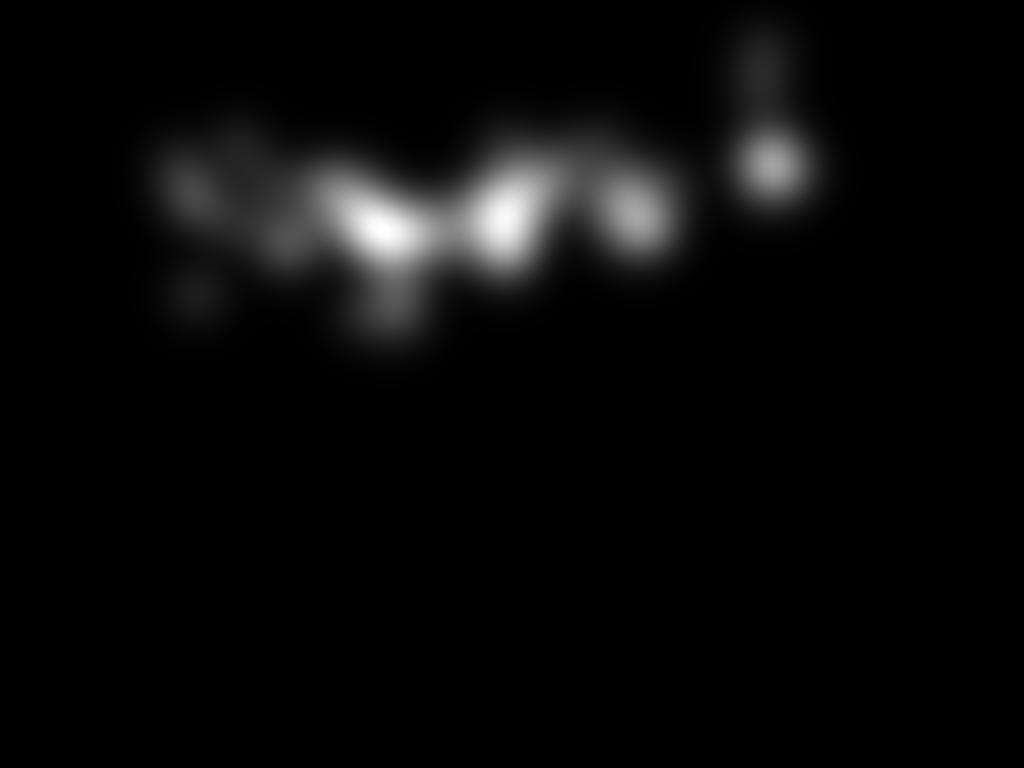}
		\end{subfigure}
		\begin{subfigure}{0.15\textwidth}
			\includegraphics[width=\textwidth]{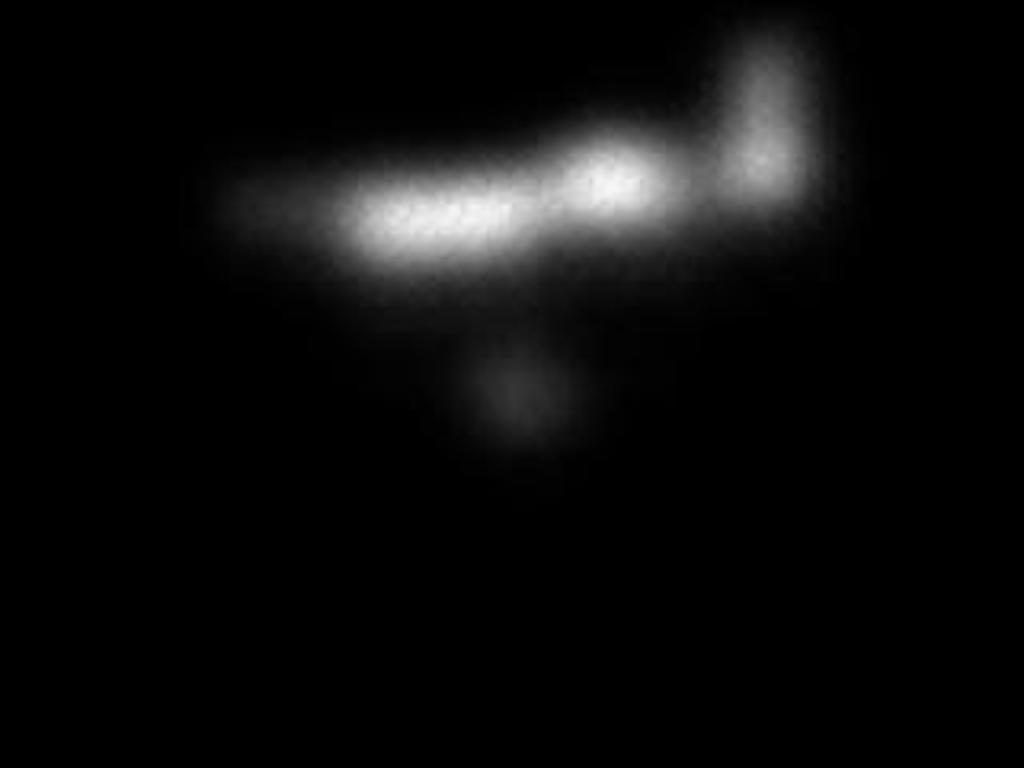}
		\end{subfigure}
		\begin{subfigure}{0.15\textwidth}
			\includegraphics[width=\textwidth]{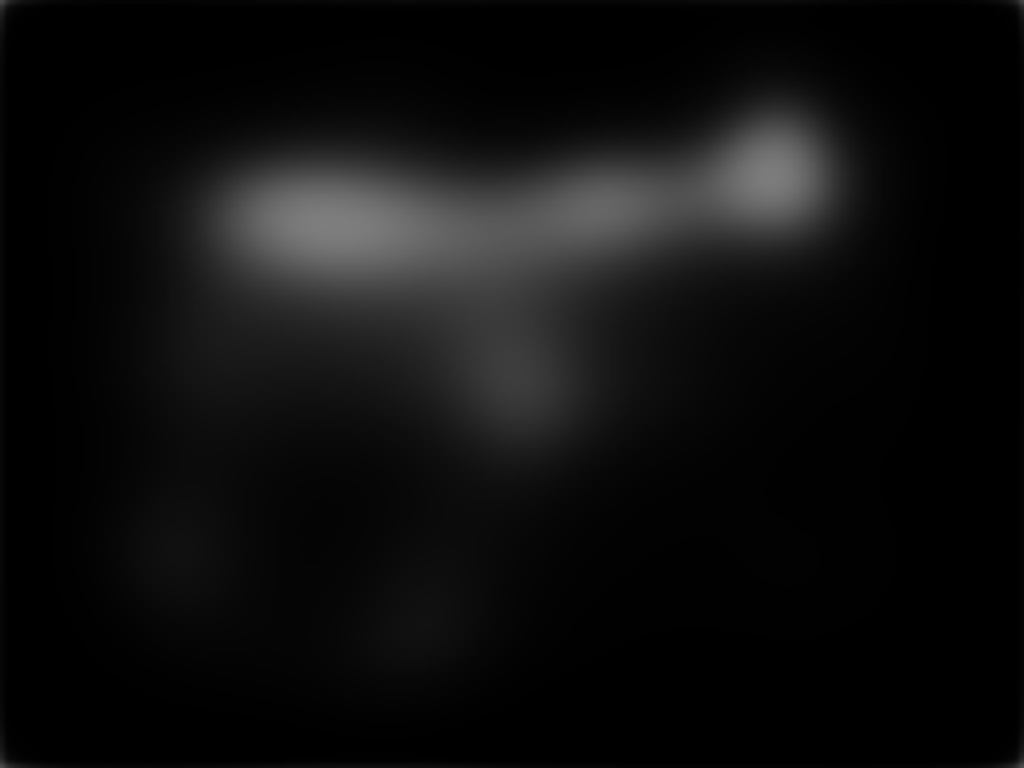}
		\end{subfigure}
		\begin{subfigure}{0.15\textwidth}
			\includegraphics[width=\textwidth]{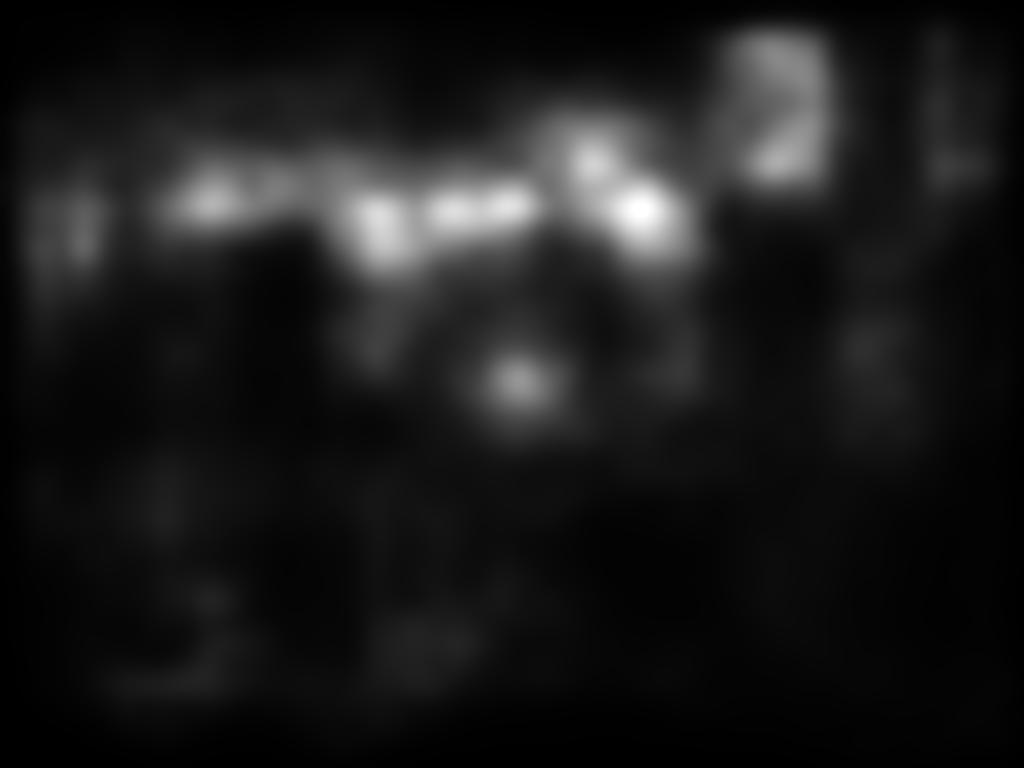}
		\end{subfigure}
		\begin{subfigure}{0.15\textwidth}
			\includegraphics[width=\textwidth]{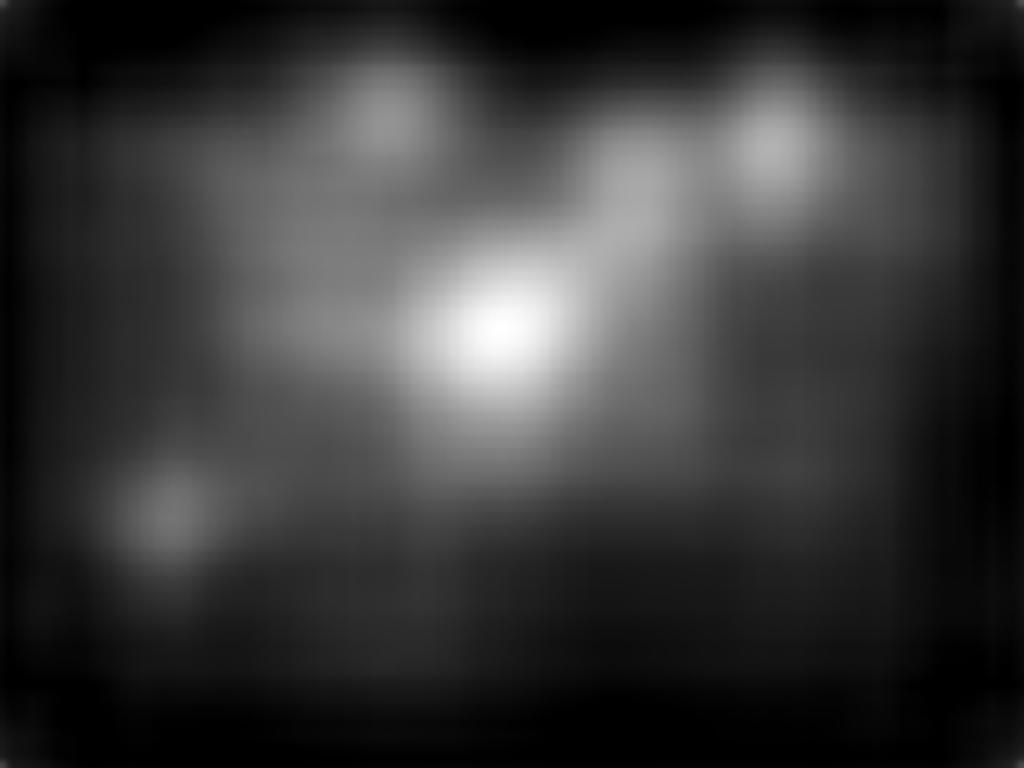}
		\end{subfigure}
		\begin{subfigure}{0.15\textwidth}
			\includegraphics[width=\textwidth]{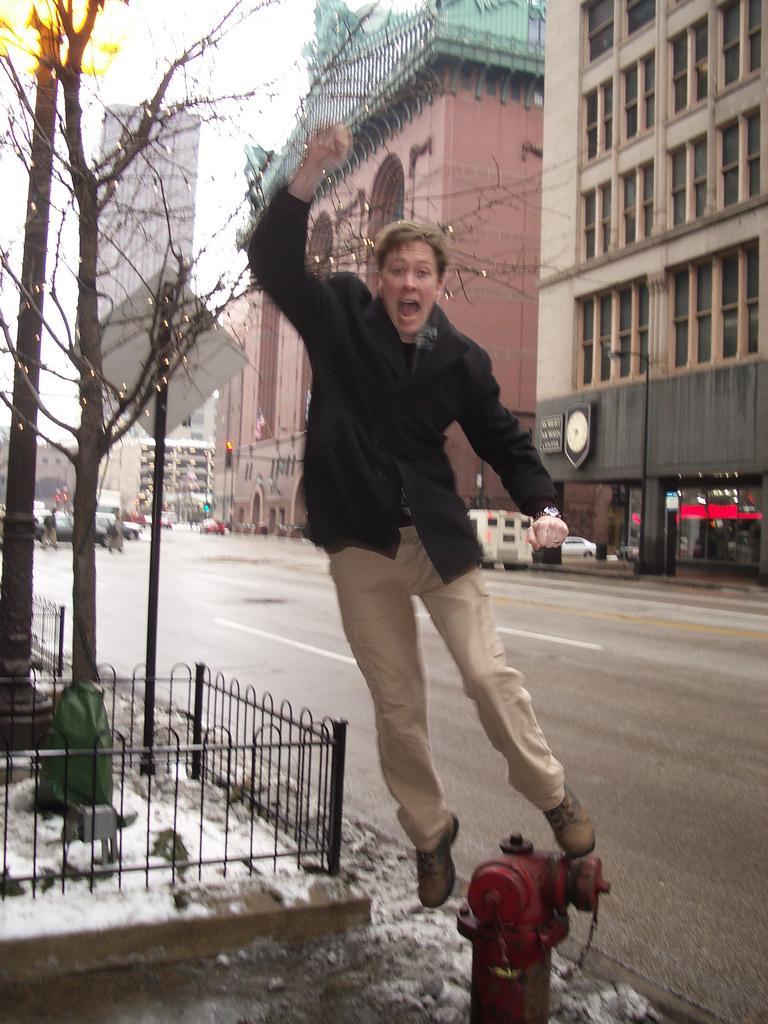}
		\end{subfigure}
		\begin{subfigure}{0.15\textwidth}
			\includegraphics[width=\textwidth]{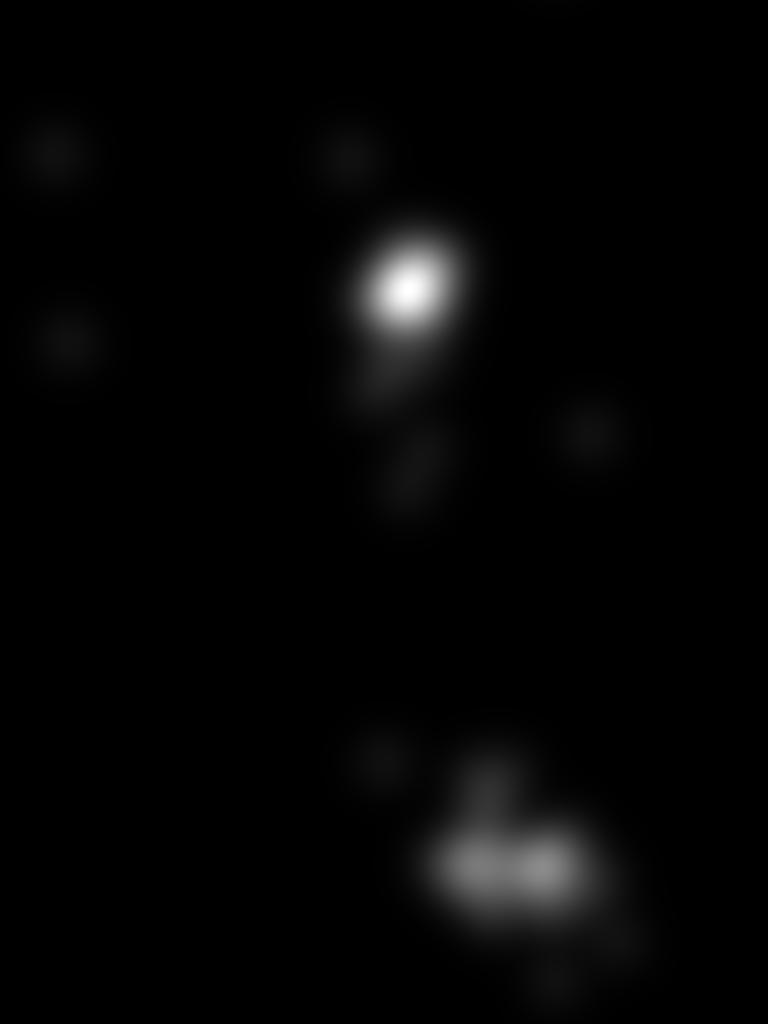}
		\end{subfigure}
		\begin{subfigure}{0.15\textwidth}
			\includegraphics[width=\textwidth]{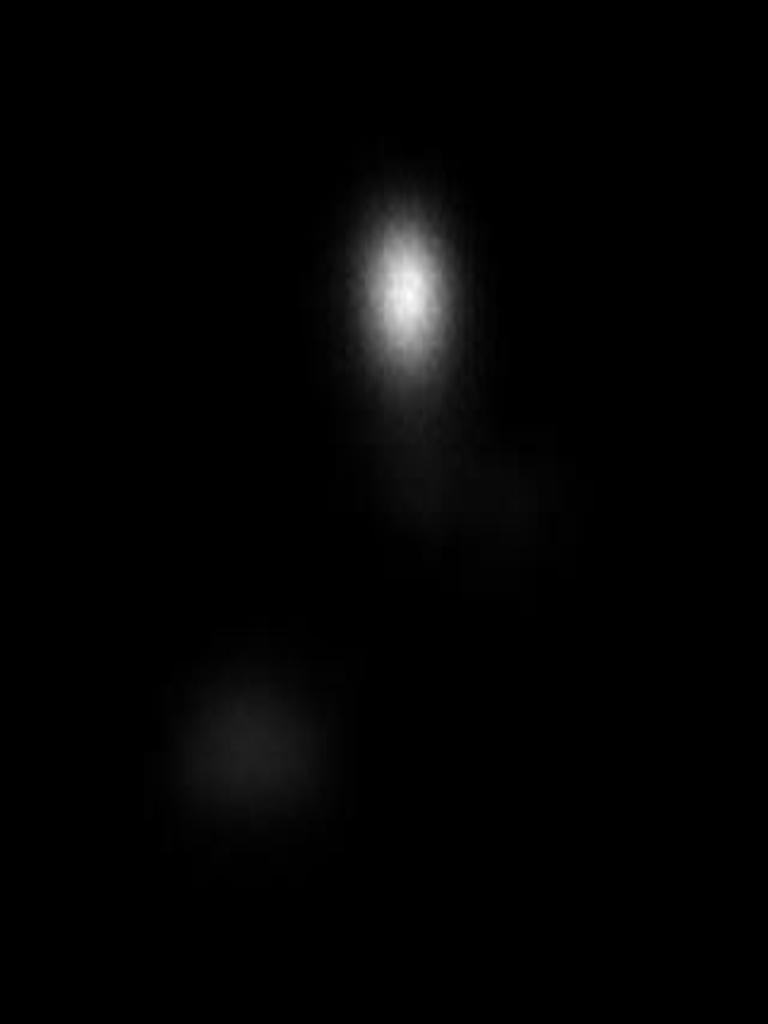}
		\end{subfigure}
		\begin{subfigure}{0.15\textwidth}
			\includegraphics[width=\textwidth]{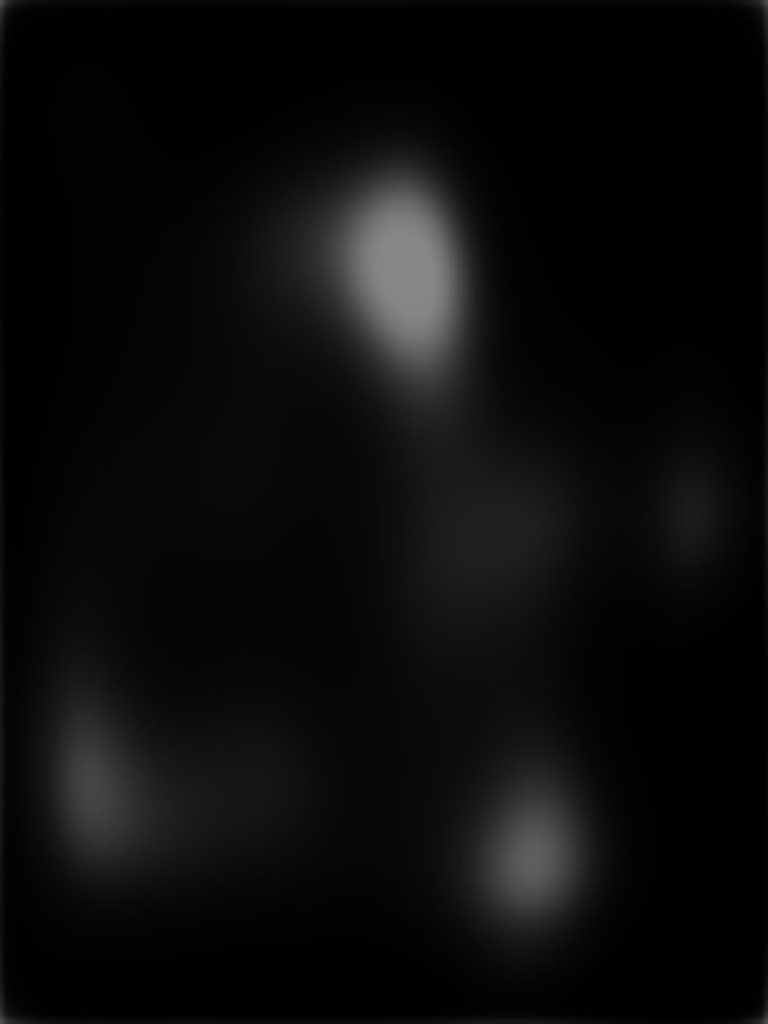}
		\end{subfigure}
		\begin{subfigure}{0.15\textwidth}
			\includegraphics[width=\textwidth]{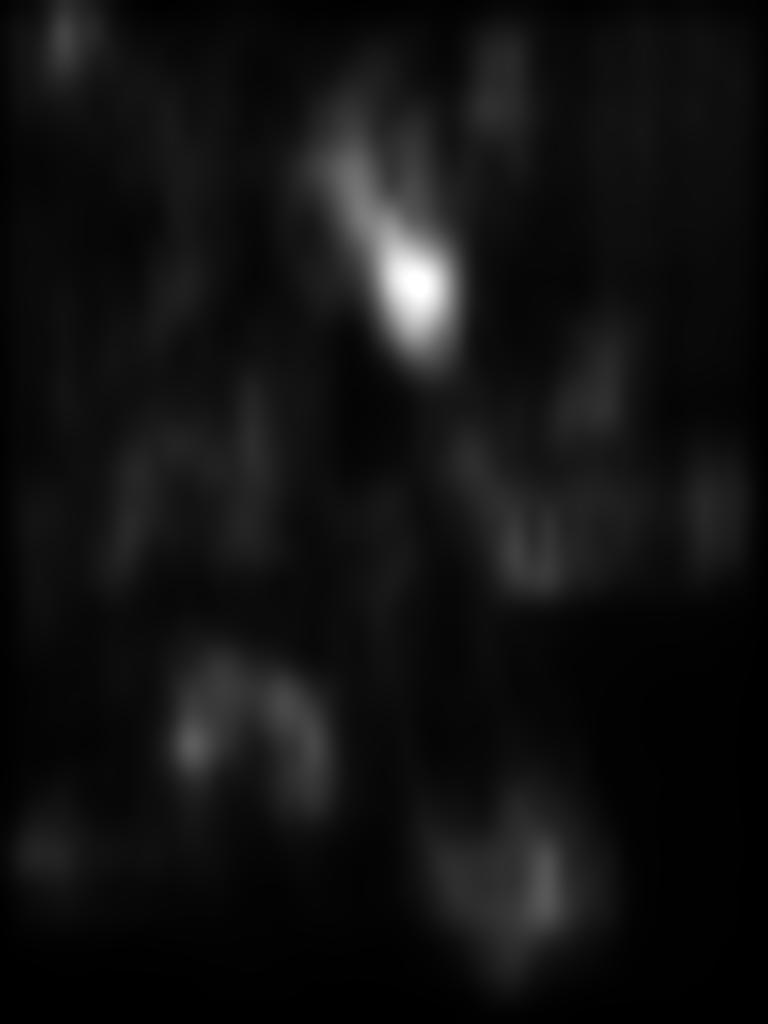}
		\end{subfigure}
		\begin{subfigure}{0.15\textwidth}
			\includegraphics[width=\textwidth]{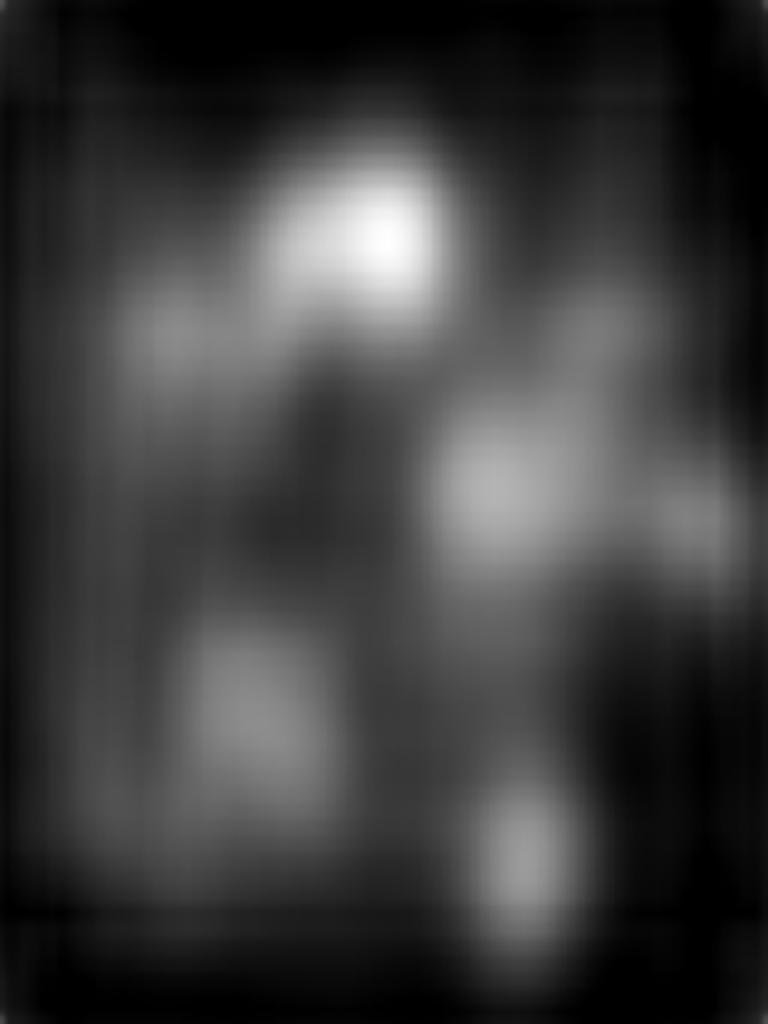}
		\end{subfigure}
		\begin{subfigure}{0.15\textwidth}
			\includegraphics[width=\textwidth]{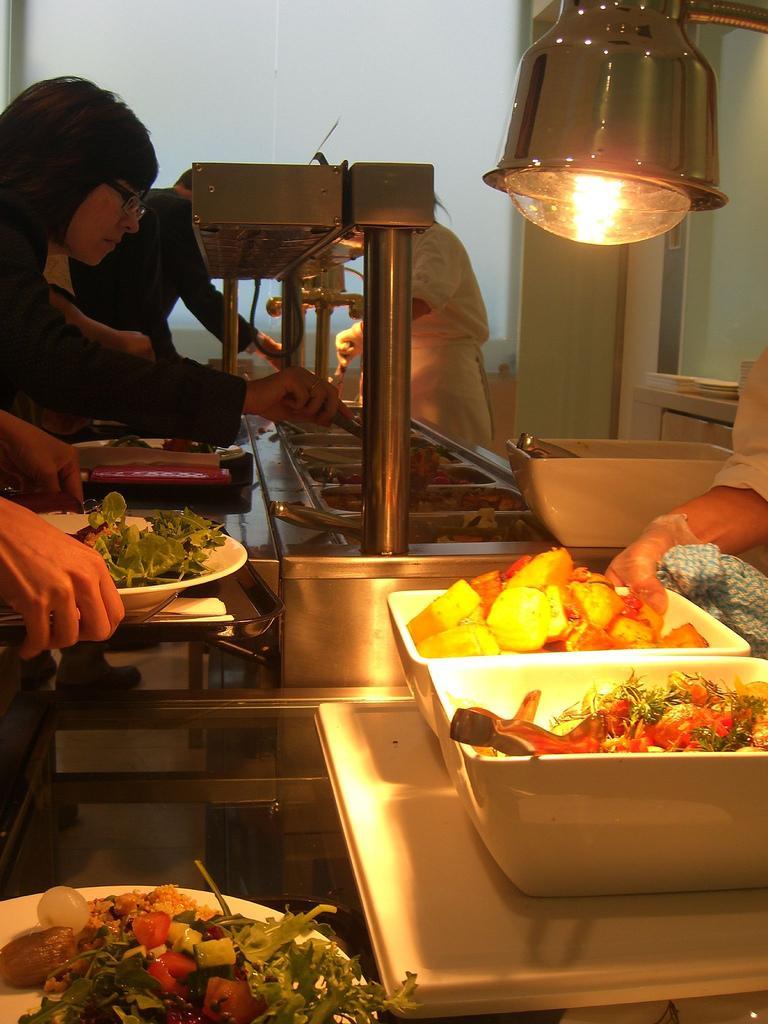}
			\caption*{Input}
		\end{subfigure}
		\begin{subfigure}{0.15\textwidth}
			\includegraphics[width=\textwidth]{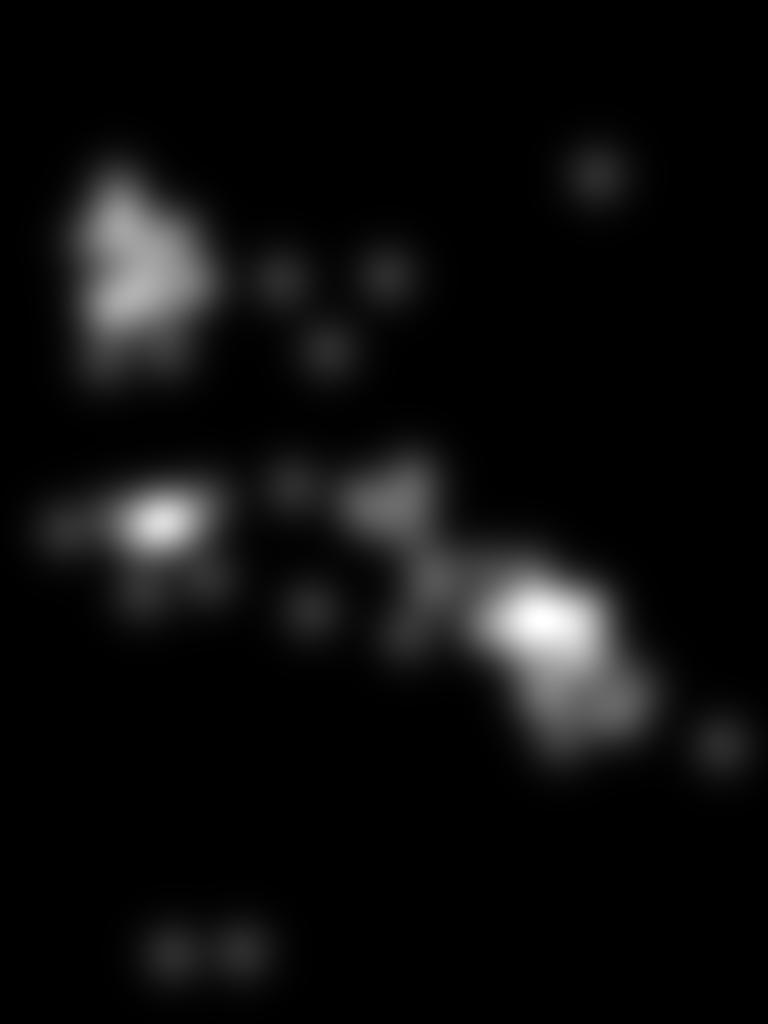}
			\caption*{GT}
		\end{subfigure}
		\begin{subfigure}{0.15\textwidth}
			\includegraphics[width=\textwidth]{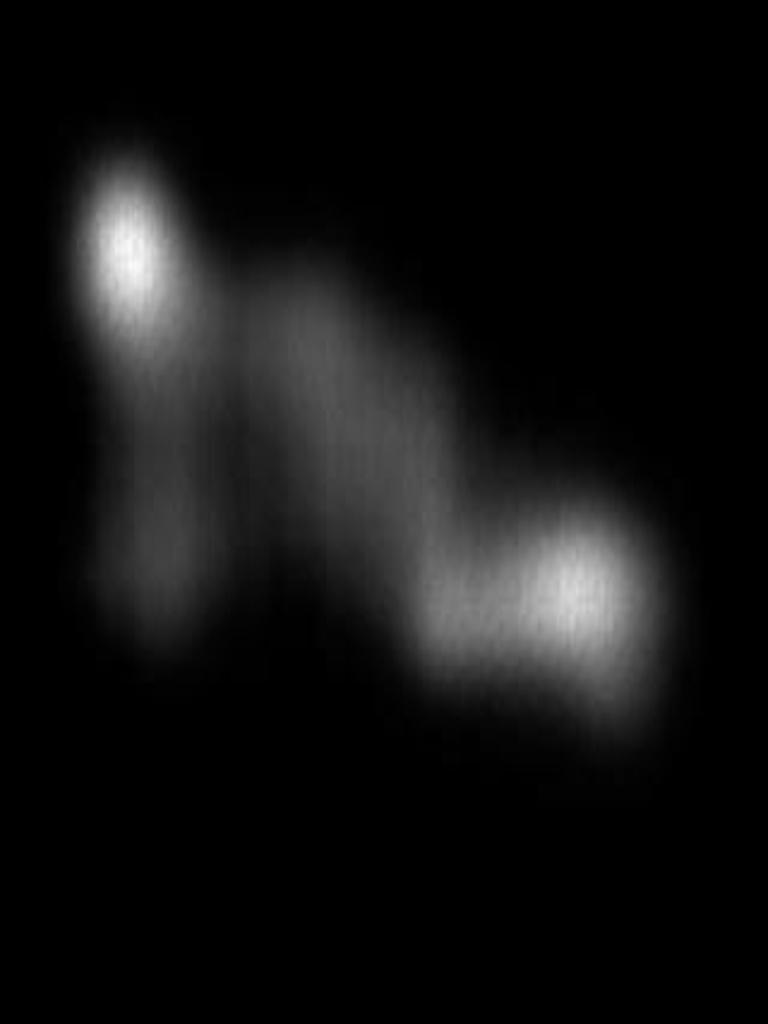}
			\caption*{Ours}
		\end{subfigure}
		\begin{subfigure}{0.15\textwidth}
			\includegraphics[width=\textwidth]{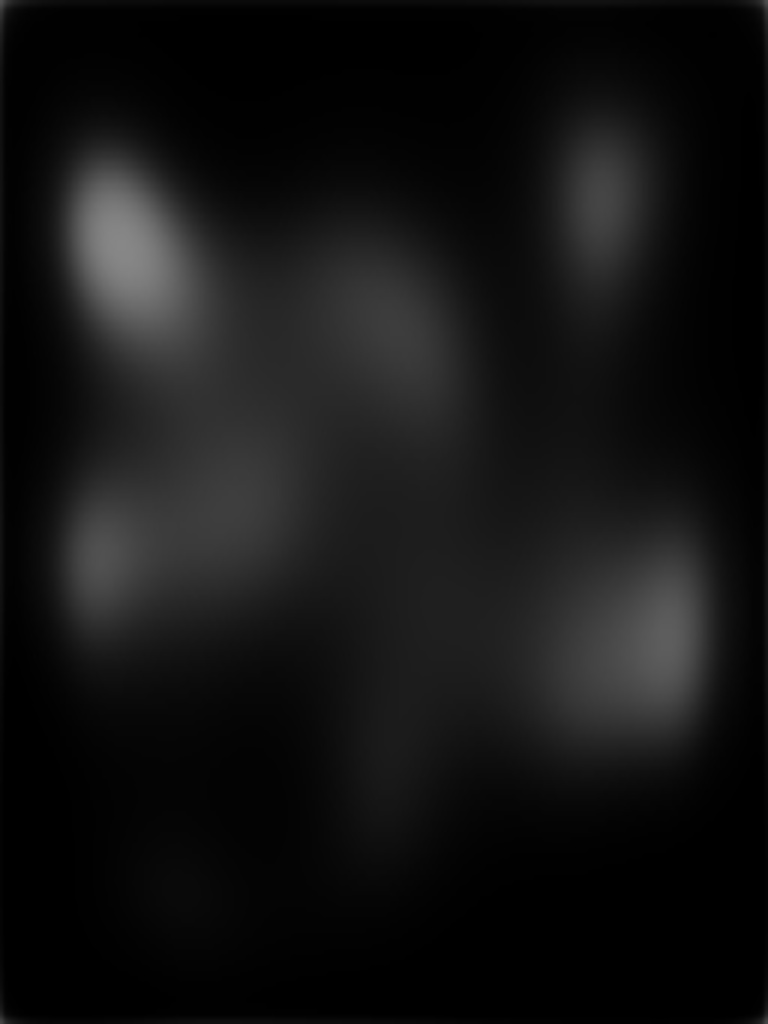}
			\caption*{SalGAN}
		\end{subfigure}
		\begin{subfigure}{0.15\textwidth}
			\includegraphics[width=\textwidth]{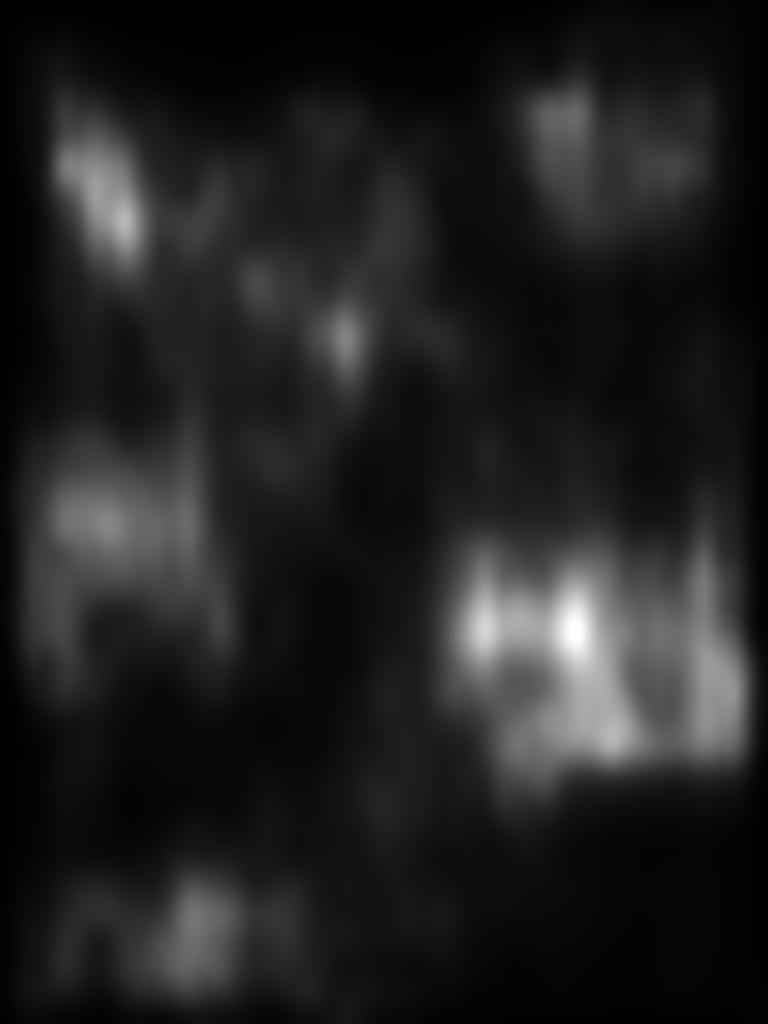}
			\caption*{Salicon*}
		\end{subfigure}
		\begin{subfigure}{0.15\textwidth}
			\includegraphics[width=\textwidth]{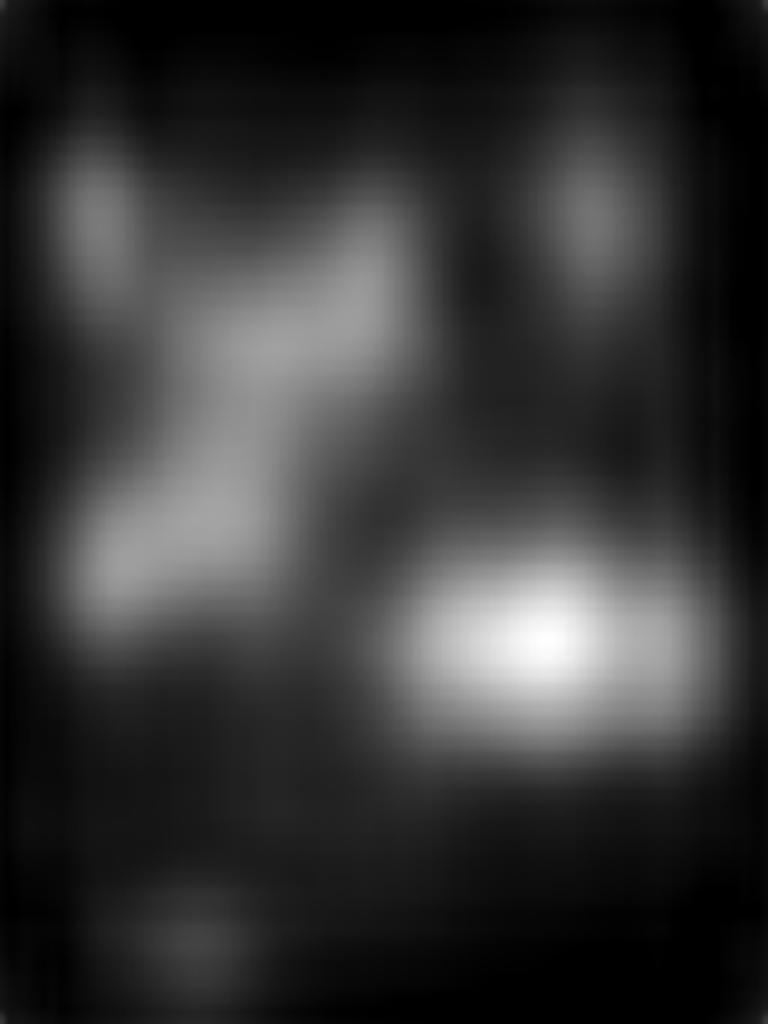}
			\caption*{Deepnet}
		\end{subfigure}
	\end{center}
    \vspace*{-15pt}
	\caption{Qualitative comparisons of different models (GT stands for Ground Truth). *As the code for Salicon is not available, we use the open source implementation \cite{thomas2016opensalicon}.}
	\label{fig:m_exa}
\end{figure*}

Tables~\ref{tab:com4} and~\ref{tab:com3} compare our model's performance with state-of-the-art models. One can see that our model performs on par or better than all single architecture models (Table~\ref{tab:modelm}), especially when considering the NSS score, which is the metric of choice for  ranking saliency models \cite{mit-saliency-benchmark}. In Table~\ref{tab:com3}, we see that our model's performance come close to considerably more complex, multi-architecture approaches such as Deepfix~\cite{kruthiventi2017deepfix} and DSCLRCN~\cite{liu2016deep}.
Figure~\ref{fig:m_exa} is a qualitative comparison of the saliency predicted by different models on some example images.

\begin{table}
\centering
\begin{tabular}{|c|c|c|c|c|}
\hline
 Model                       &NSS        &CC                       &AUC        &Sim\\
\hline
\hline
Salicon*~\cite{huang2015salicon}          &1.557        &0.659                     &0.808                  &0.600\\
SalGAN~\cite{pan2017salgan}             &1.816        &0.844                     &\textbf{0.857}                  &0.728\\
Deepnet~\cite{pan2016shallow}                  &1.555         &0.763                     &0.840                  &0.639\\
\hline
proposed                  &\textbf{1.896}         &\textbf{0.871}                     &0.852                  &\textbf{0.760}\\
\hline
\end{tabular}
\caption{Comparison of different models on LSUN 2017 saliency prediction challenge \cite{WinNT} (SALICON testing dataset). *As the code for Salicon is not available, we use the open source implementation  \cite{thomas2016opensalicon}.}
\label{tab:com4}
\end{table} 

\begin{table}
\centering
\begin{tabular}{|c|c|c|c|c|}
\hline
 Model                       &NSS        &CC                       &AUC        &Sim\\
\hline
\hline
DSCLRCN~\cite{liu2016deep}       &\textbf{2.35}          &\textbf{0.80}                    &\textbf{0.87}                  &\textbf{0.68}\\
Deepfix~\cite{kruthiventi2017deepfix}          &2.26         &0.78                    &\textbf{0.87}                  &0.67\\
\hline
\textit{proposed}                  &\textit{2.17 }        &\textit{0.74 }                    &\textit{0.83}                  &\textit{0.60}\\
\hline
Salicon~\cite{huang2015salicon}          &2.12        &0.74                     &\textbf{0.87}                  &0.67\\
SalGAN~\cite{pan2017salgan}             &2.04        &0.73                     &0.86                  &0.63\\
PDP~\cite{jetley2016end}                  &2.05         &0.70                     &0.85                  &0.60\\
DVA~\cite{wang2017deep}                   &1.98          &0.68                    &0.85                   &0.58\\
\hline
\end{tabular}
\caption{The comparison of different models on MIT300 dataset.}
\label{tab:com3}
\end{table} 

\section{Visualizing salient patterns}
This section analyses what is learnt by state-of-the-art deep saliency models using the visualization tools discussed in section~\ref{sec:viz}.
The model proposed in section~\ref{sec:model} is trained on the SALICON training dataset as before, but without fine-tuning it on MIT1003 dataset to avoid overfitting when computing the NSS score for each activation map. 
In addition to the proposed model, we apply a similar analysis on three deep saliency models for which the code is publicly available: Deepnet, SalGAN and OpenSalicon. In Deepnet \cite{pan2016shallow}, the first five convolutional layers were determined as the feature extraction part. In SalGAN \cite{pan2017salgan}, the encoder was treated as the feature extraction part. In OpenSalicon~\cite{thomas2016opensalicon}, both coarse scale (Saliconc) and fine scale (Saliconf) were visualized.

\begin{figure*}
	\begin{center}
		\begin{subfigure}{0.18\textwidth}
			\includegraphics[width=\textwidth]{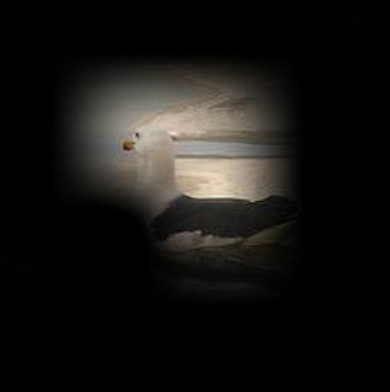}
		\end{subfigure}
		\begin{subfigure}{0.18\textwidth}
			\includegraphics[width=\textwidth]{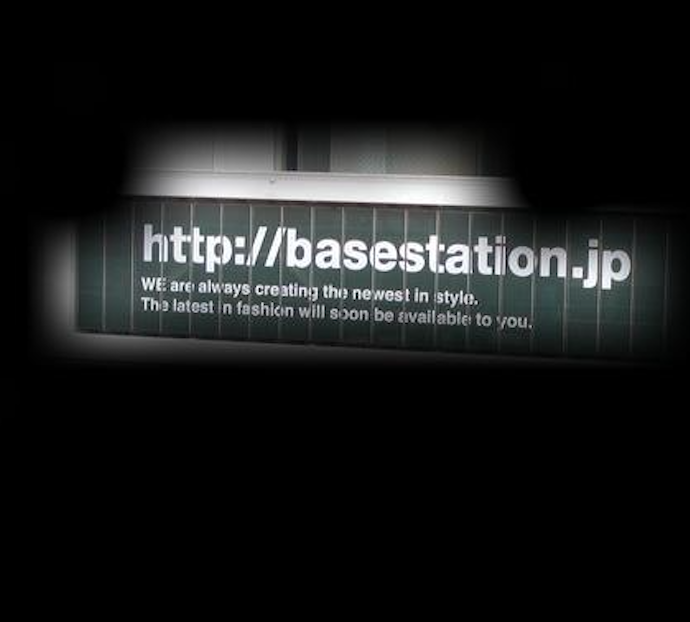}
		\end{subfigure}
		\begin{subfigure}{0.18\textwidth}
			\includegraphics[width=\textwidth]{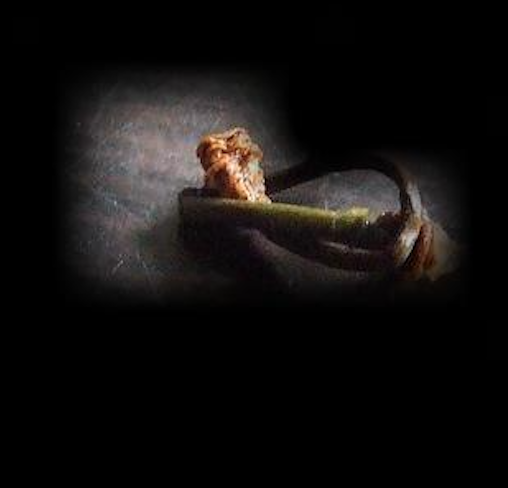}
		\end{subfigure}
		\begin{subfigure}{0.18\textwidth}
			\includegraphics[width=\textwidth]{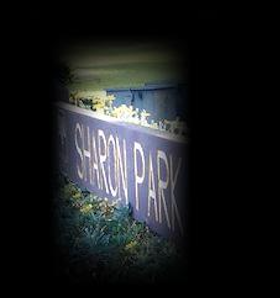}
		\end{subfigure}
                 \begin{subfigure}{0.18\textwidth}
			\includegraphics[width=\textwidth]{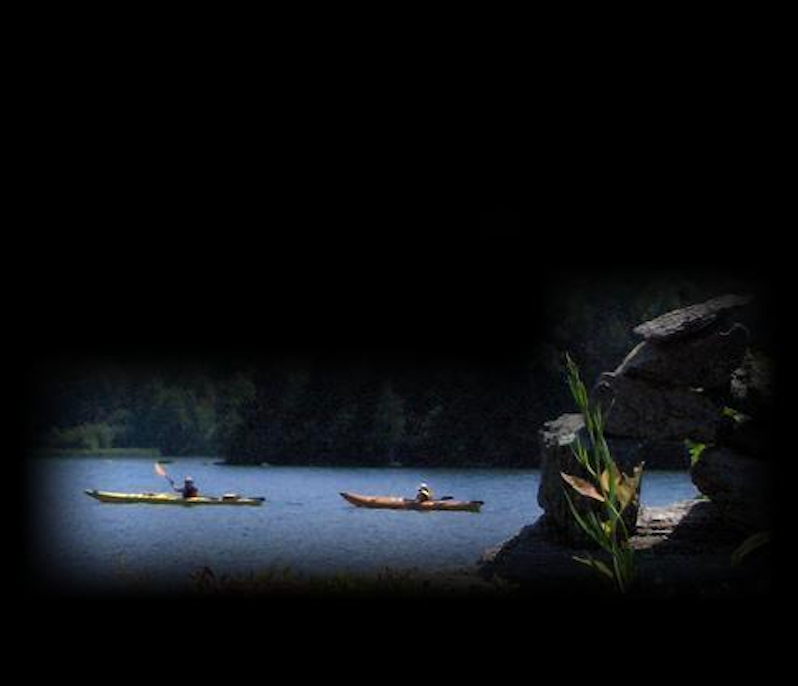}
		\end{subfigure}
	\end{center}
    \vspace*{-15pt}
	\caption{Examples of patterns produced for the activation map 115 of the proposed model, with a mean NSS score of 4.5808.}
	\label{fig:h_our}
\end{figure*}

\begin{figure*}
	\begin{center}
		\begin{subfigure}{0.18\textwidth}
			\includegraphics[width=\textwidth]{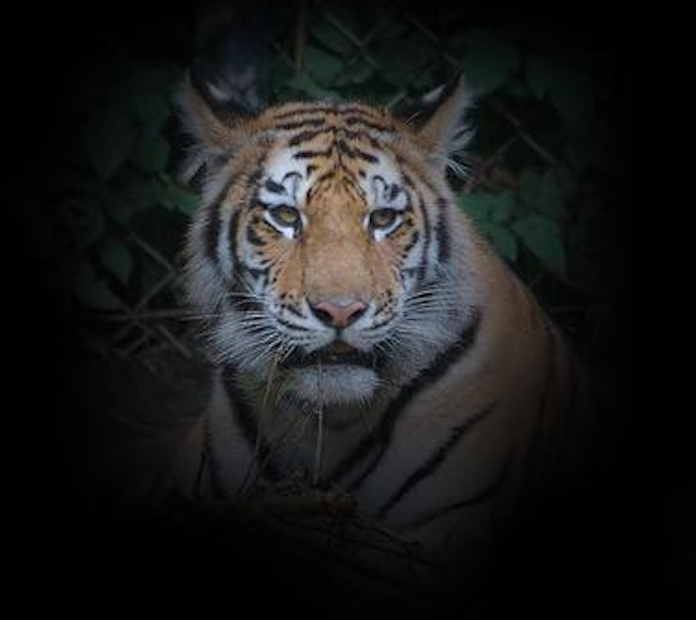}
		\end{subfigure}
		\begin{subfigure}{0.18\textwidth}
			\includegraphics[width=\textwidth]{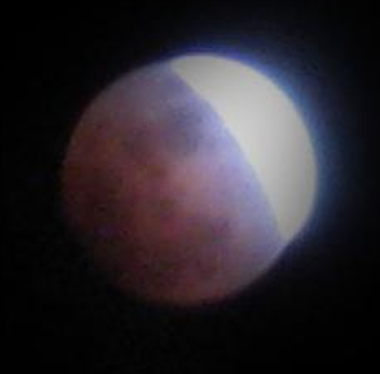}
		\end{subfigure}
		\begin{subfigure}{0.18\textwidth}
			\includegraphics[width=\textwidth]{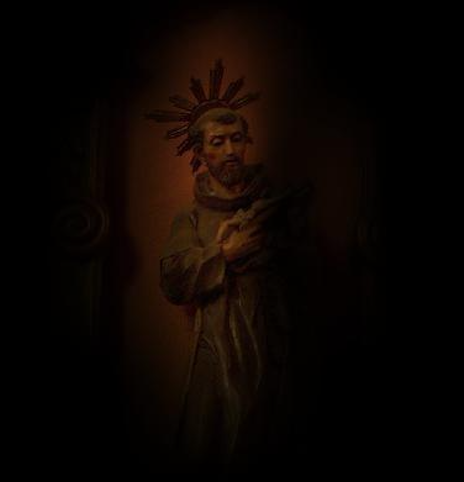}
		\end{subfigure}
		\begin{subfigure}{0.18\textwidth}
			\includegraphics[width=\textwidth]{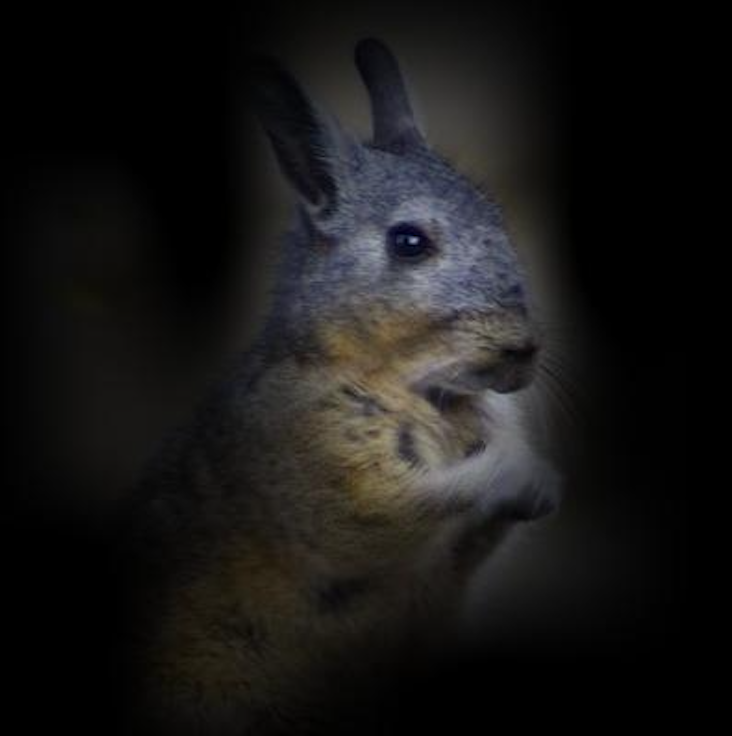}
		\end{subfigure}
                 \begin{subfigure}{0.18\textwidth}
			\includegraphics[width=\textwidth]{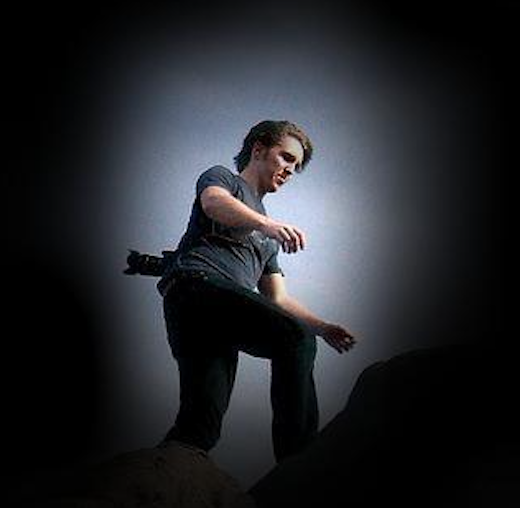}
		\end{subfigure}
	\end{center}
    \vspace*{-15pt}
	\caption{Example of patterns produced for the activation map 221 of SalGAN, with a mean NSS score of 4.6019.}
	\label{fig:h_salgan}
\end{figure*}

\begin{figure*}
	\begin{center}
		\begin{subfigure}{0.18\textwidth}
			\includegraphics[width=\textwidth]{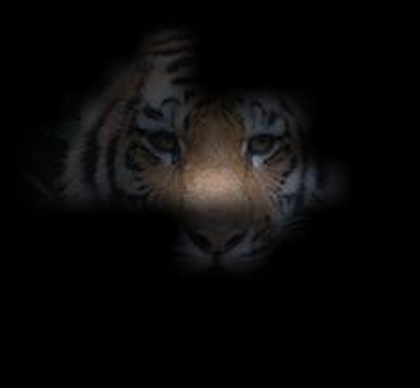}
		\end{subfigure}
		\begin{subfigure}{0.18\textwidth}
			\includegraphics[width=\textwidth]{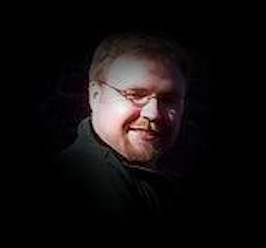}
		\end{subfigure}
		\begin{subfigure}{0.18\textwidth}
			\includegraphics[width=\textwidth]{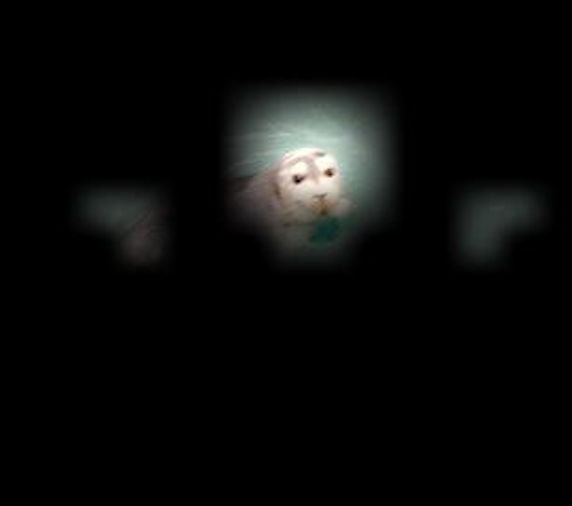}
		\end{subfigure}
		\begin{subfigure}{0.18\textwidth}
			\includegraphics[width=\textwidth]{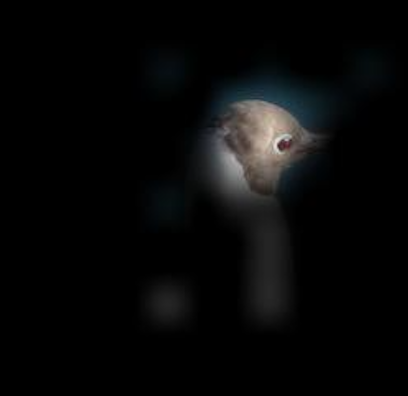}
		\end{subfigure}
                 \begin{subfigure}{0.18\textwidth}
			\includegraphics[width=\textwidth]{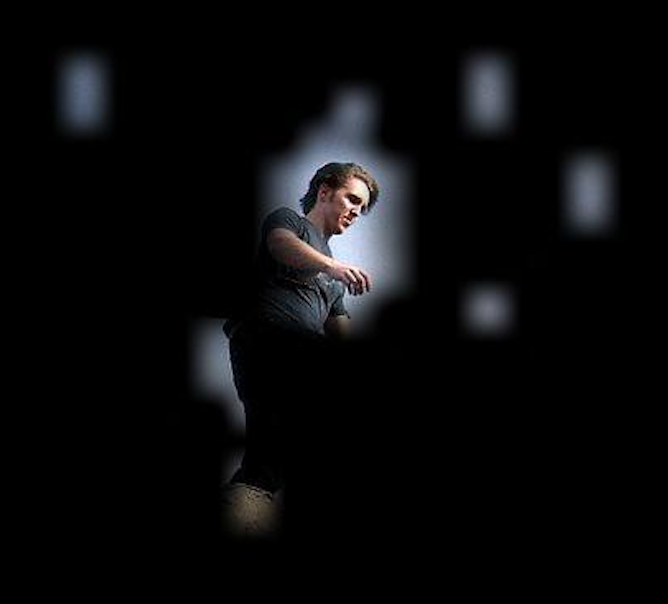}
		\end{subfigure}
	\end{center}
    \vspace*{-15pt}
	\caption{Example of patterns produced for the activation map 434 of Salicon at fine resolution (Saliconf), with a mean NSS score of 5.3637.}
	\label{fig:h_open_h}
\end{figure*}

\begin{figure*}
	\begin{center}
		\begin{subfigure}{0.18\textwidth}
			\includegraphics[width=\textwidth]{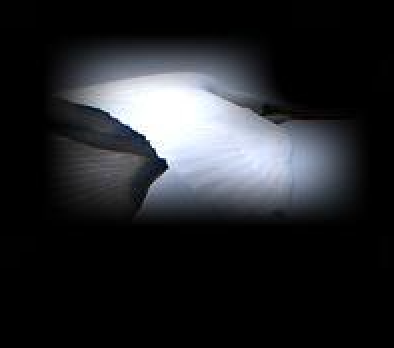}
		\end{subfigure}
		\begin{subfigure}{0.18\textwidth}
			\includegraphics[width=\textwidth]{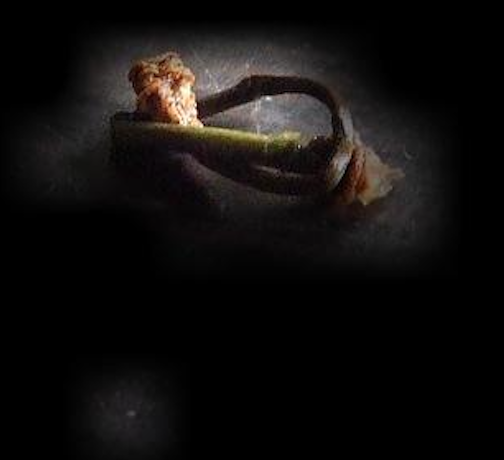}
		\end{subfigure}
		\begin{subfigure}{0.18\textwidth}
			\includegraphics[width=\textwidth]{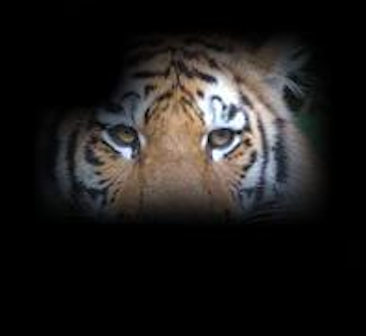}
		\end{subfigure}
		\begin{subfigure}{0.18\textwidth}
			\includegraphics[width=\textwidth]{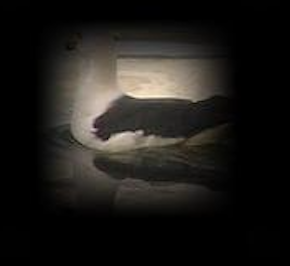}
		\end{subfigure}
                 \begin{subfigure}{0.18\textwidth}
			\includegraphics[width=\textwidth]{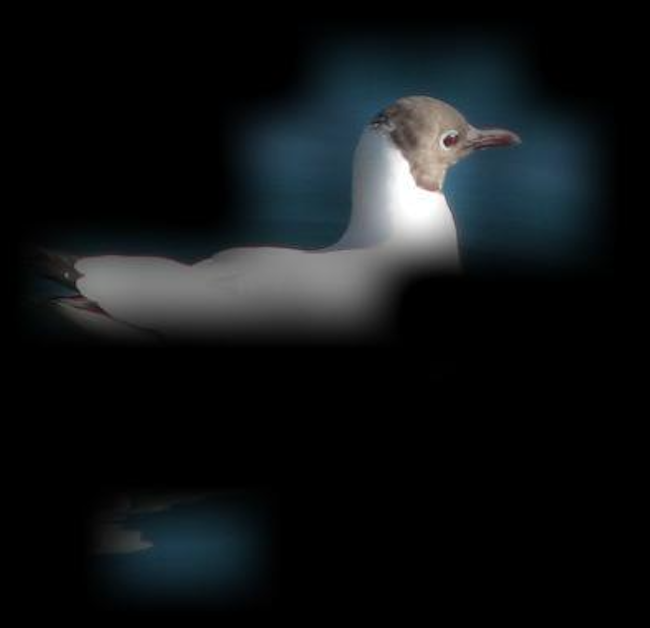}
		\end{subfigure}
	\end{center}
    \vspace*{-15pt}
	\caption{Example of patterns produced for the activation map 232 of Salicon at coarse resolution (Saliconc), with a mean NSS score of 5.0027.}
	\label{fig:c_open_h}
\end{figure*}

\begin{figure*}
	\begin{center}
		\begin{subfigure}{0.18\textwidth}
			\includegraphics[width=\textwidth]{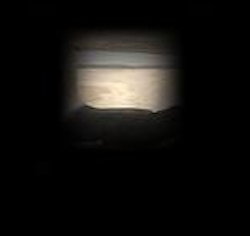}
                \vspace*{-10pt}
		\end{subfigure}
		\begin{subfigure}{0.18\textwidth}
			\includegraphics[width=\textwidth]{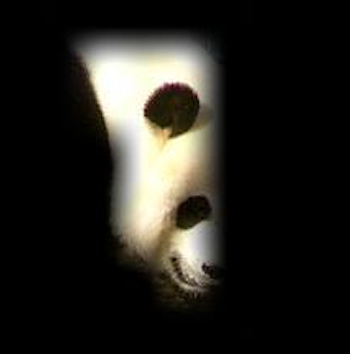}
            \vspace*{-10pt}
		\end{subfigure}
		\begin{subfigure}{0.18\textwidth}
			\includegraphics[width=\textwidth]{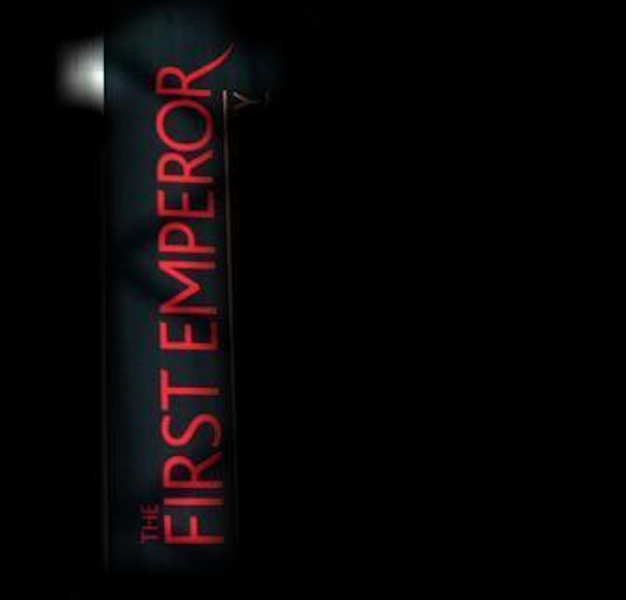}
            \vspace*{-10pt}
		\end{subfigure}
		\begin{subfigure}{0.18\textwidth}
			\includegraphics[width=\textwidth]{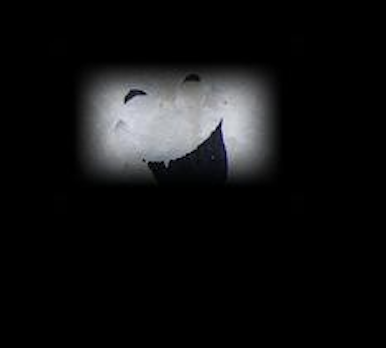}
            \vspace*{-10pt}
		\end{subfigure}
                 \begin{subfigure}{0.18\textwidth}
			\includegraphics[width=\textwidth]{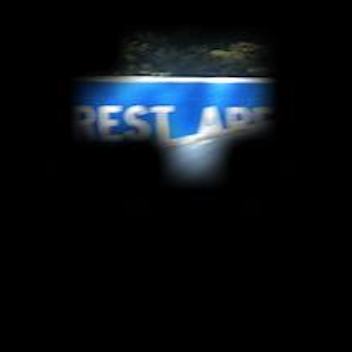}
            \vspace*{-10pt}
		\end{subfigure}
	\end{center}
    \vspace*{-15pt}
	\caption{Example of patterns produced for the activation map 162 of Deepnet, with a mean NSS score of 4.0101.}
	\label{fig:deepnet_h}
\end{figure*}

Figures~\ref{fig:h_our} to~\ref{fig:deepnet_h} show example patterns for  activation maps with high mean NSS scores (far beyond the model's performance) in different models. The patterns in these figures are generated as the product of an input image with the activation map, cropped to the active areas for legibility. From those figures, we can see that most activation maps with high mean NSS score focus on \textit{a unique object or part of an object} (head or face, etc). 

\begin{figure*}
	\begin{center}
		\begin{subfigure}{0.18\textwidth}
			\includegraphics[width=\textwidth]{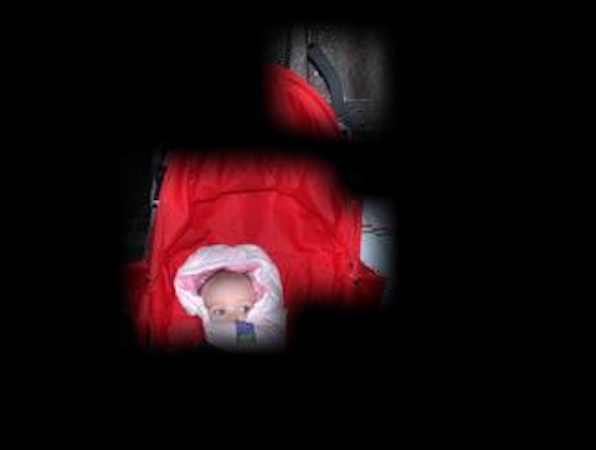}
		        \caption{Deepnet}
		\end{subfigure}
		\begin{subfigure}{0.18\textwidth}
			\includegraphics[width=\textwidth]{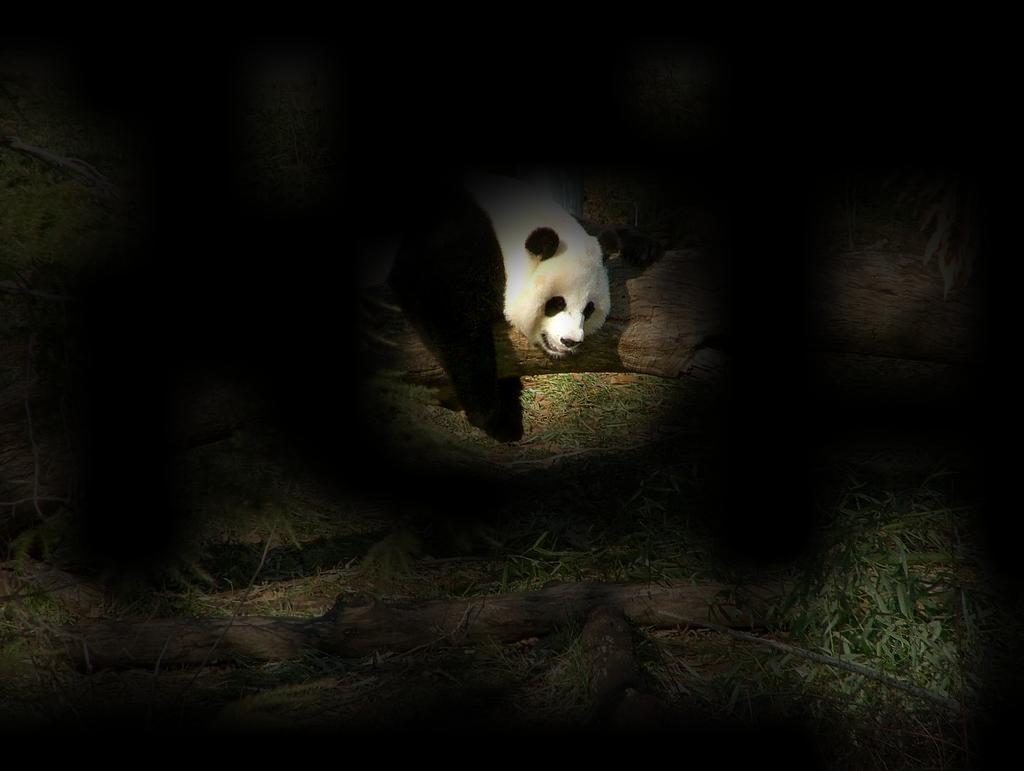}
			\caption{SalGAN}
		\end{subfigure}
		\begin{subfigure}{0.18\textwidth}
			\includegraphics[width=\textwidth]{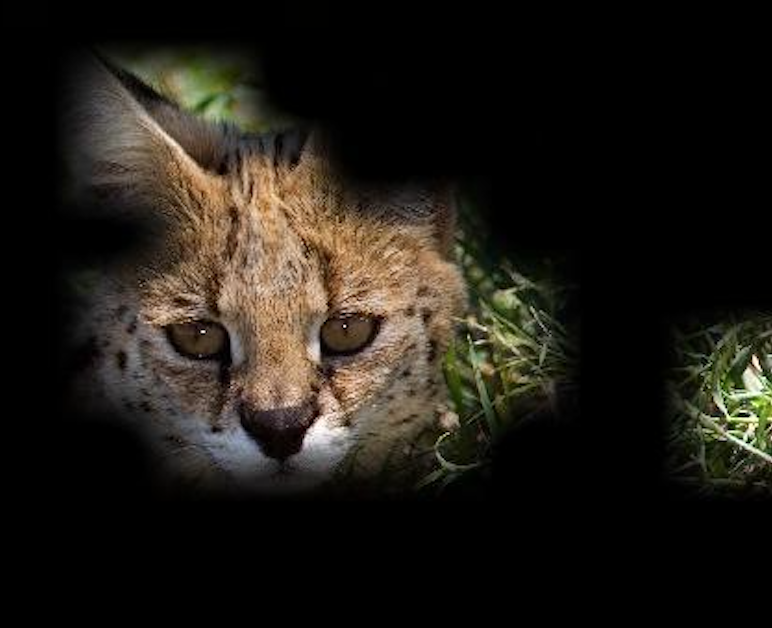}
			\caption{Saliconc}
		\end{subfigure}
		\begin{subfigure}{0.18\textwidth}
			\includegraphics[width=\textwidth]{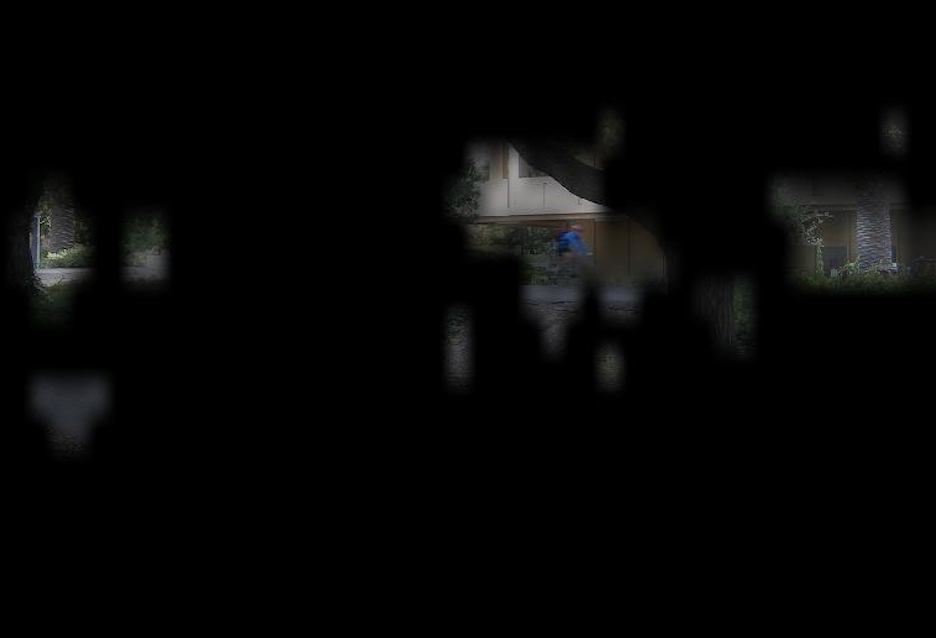}
			\caption{Saliconf}
		\end{subfigure}
                 \begin{subfigure}{0.18\textwidth}
			\includegraphics[width=\textwidth]{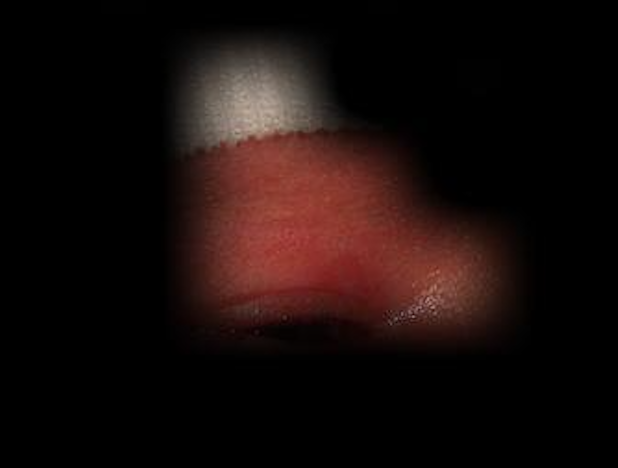}
			\caption{ours}
		\end{subfigure}
	\end{center}
    \vspace*{-15pt}
	\caption{Example patterns for activation maps with medium mean NSS score, drawn for different models.}
	\label{fig:all_m}
\end{figure*}

\begin{figure*}
\vspace*{-10pt}
	\begin{center}
		\begin{subfigure}{0.18\textwidth}
			\includegraphics[width=\textwidth]{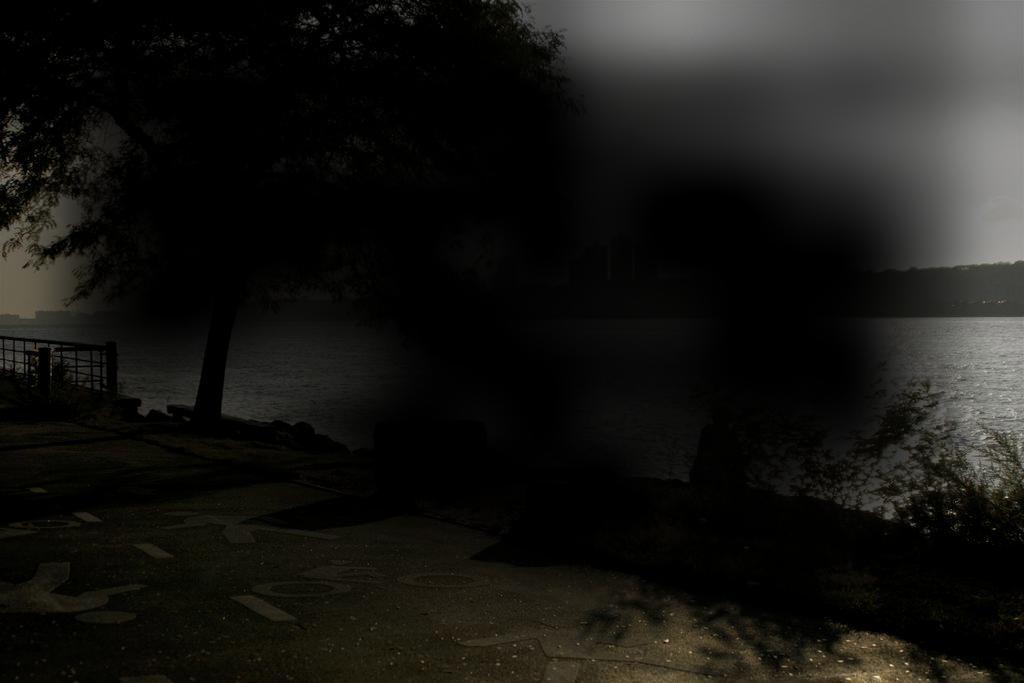}
		        \caption{Deepnet}
              
		\end{subfigure}
		\begin{subfigure}{0.18\textwidth}
			\includegraphics[width=\textwidth]{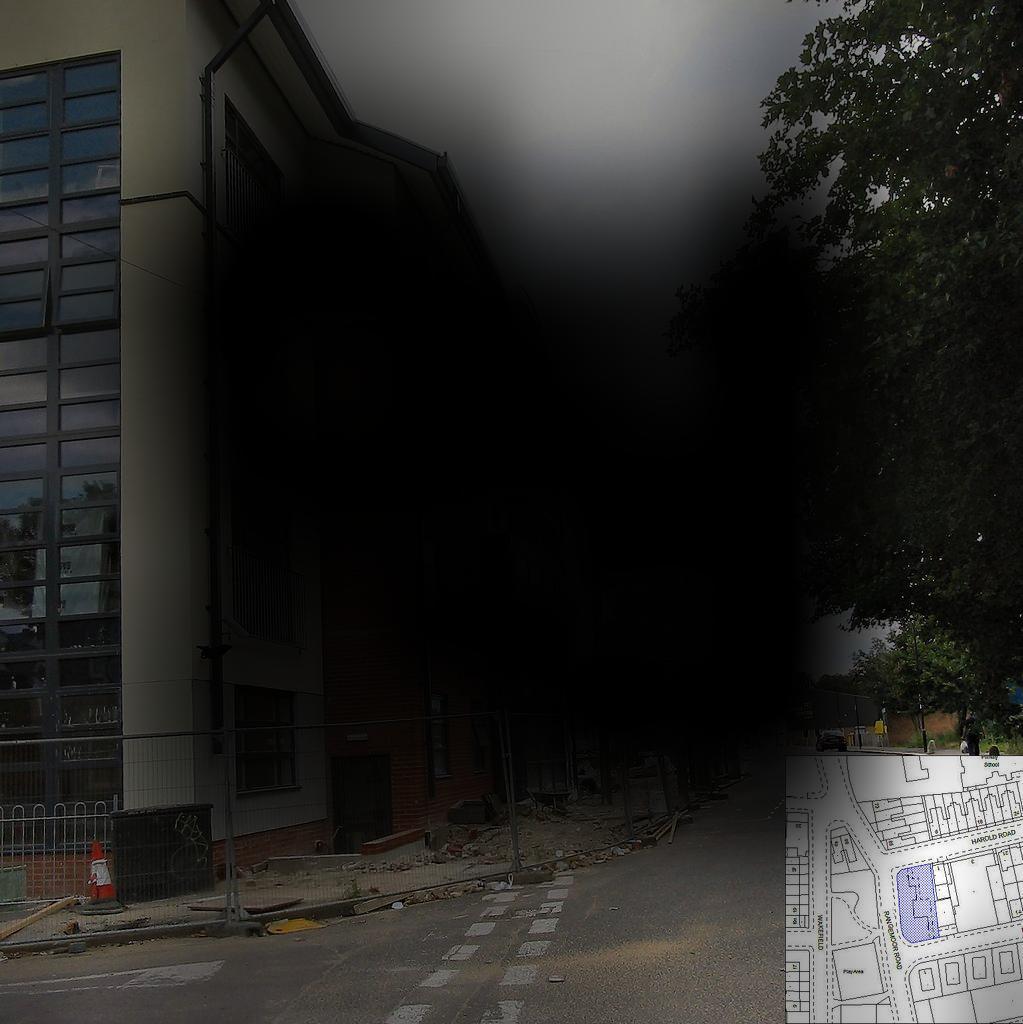}
			\caption{SalGAN}

		\end{subfigure}
		\begin{subfigure}{0.18\textwidth}
			\includegraphics[width=\textwidth]{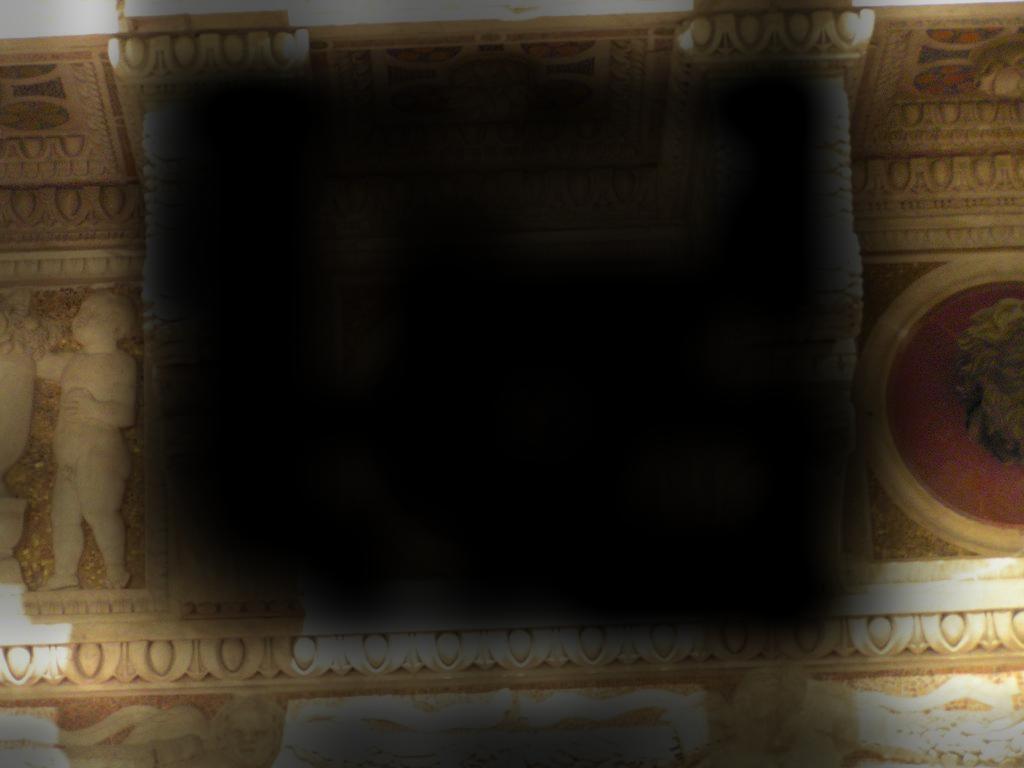}
			\caption{Saliconc}

		\end{subfigure}
		\begin{subfigure}{0.18\textwidth}
			\includegraphics[width=\textwidth]{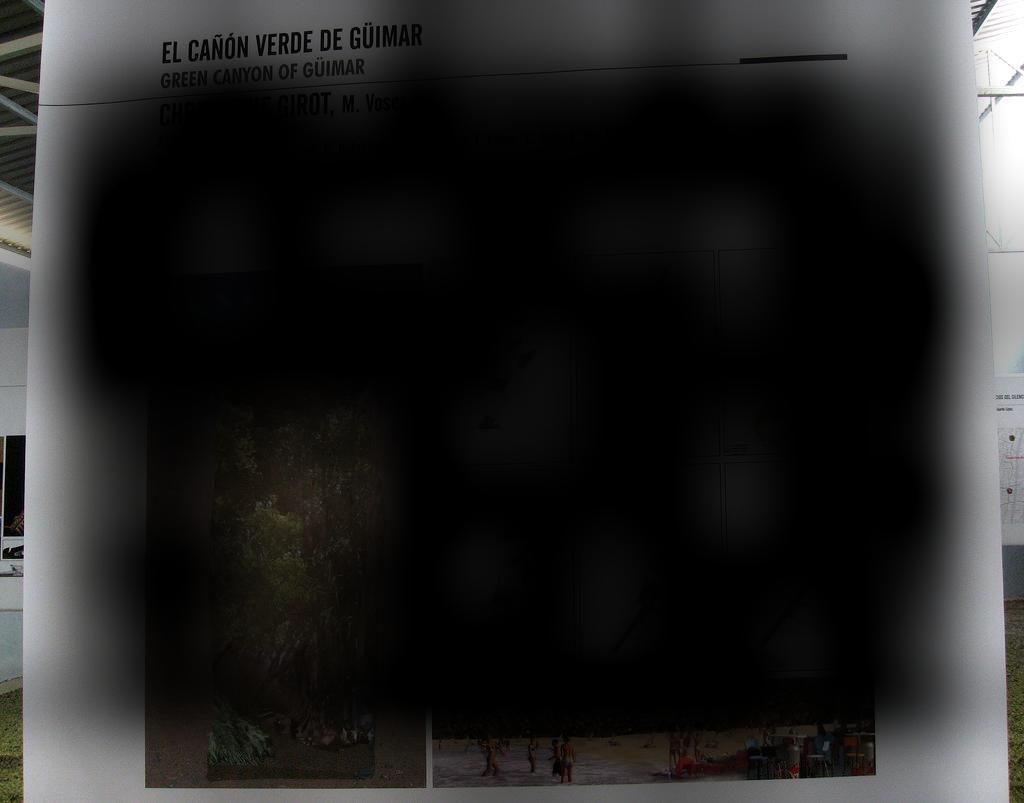}
			\caption{Saliconf}

		\end{subfigure}
                 \begin{subfigure}{0.18\textwidth}
			\includegraphics[width=\textwidth]{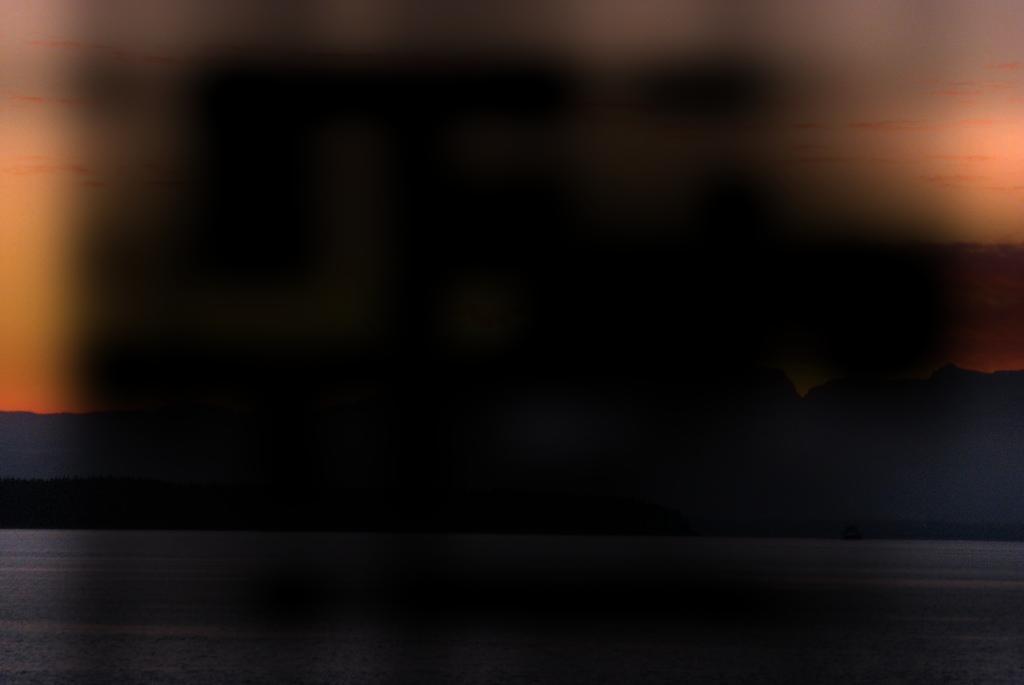}
			\caption{ours}

		\end{subfigure}
	\end{center}
    \vspace*{-15pt}
	\caption{Example patterns for activation maps with low mean NSS score, drawn for different models.}
	\label{fig:all_l}
\end{figure*}

Figure~\ref{fig:all_m} shows example patterns for activation maps with medium mean NSS scores. In these examples, we can see that the active regions are less clearly focused on a single object or part. 
More importantly, for activation maps with low mean NSS score, shown in Figure~\ref{fig:all_l}, the patterns show a negative central bias, clearly inhibiting the central part of the image. Note that the models analyzed in this figure do not include an explicit central bias constraint, therefore this bias has been learnt solely from the training data and appears to be encoded by low NSS neurons.  

Since some activation maps (positive fixation detectors) have very high mean NSS scores, we investigate the relationships between the model performance (here we use the NSS score in MIT1003 dataset as the model performance) and the proportion of positive fixation detectors. Table~\ref{tab:p_fix} records this ratio for all analyzed models. We can see that models with better overall performance have higher proportions of fixation detectors (with a correlation coefficient of $r=0.94$), and our model has the highest ratio of positive fixation detectors amongst all analyzed models. Note that in all cases, the ratio of positive fixation detectors remains small.  

\begin{table}
\centering
\begin{tabular}{|c|c|c|c|c|}
\hline
Model             & \# activation maps & \# positive detectors & ratio &NSS\\
\hline
\hline
Deepnet~\cite{pan2016shallow} &512 & 5 &  $\sim$1\%  &1.68\\
SalGAN~\cite{pan2017salgan} &512 & 14 & 2.7\% &2.15\\
OpenSalicon~\cite{huang2015salicon,thomas2016opensalicon} &1024 & 19 & 1.8\% &1.92 \\
proposed & 512 & 21 & 4.1\% &2.21 \\
\hline
\end{tabular}
\caption{The relationship between the model performance and the number of positive fixation detectors inside the model. The correlation between NSS and ratio is 0.94.}
\label{tab:p_fix}
\end{table}

After determining which activation maps are positive fixation detectors inside the deep models, the question remains of what are those detectors are attuned to (i.e., objects and object parts). For this purpose we use measure of the normalized mean detection frequency $f_n(c)$ of a class $c$ as the ratio:
\begin{align}
f_n(c) = \frac{f_d(c)}{f_t(c)}
\end{align}
where $f_t$ is the total number of occurrence of the class in the dataset and $f_d$ is the number of detected instances.

\begin{figure*}
	\begin{center}
		\begin{subfigure}{\textwidth}
        \centering
        	\includegraphics[width=0.4\textwidth]{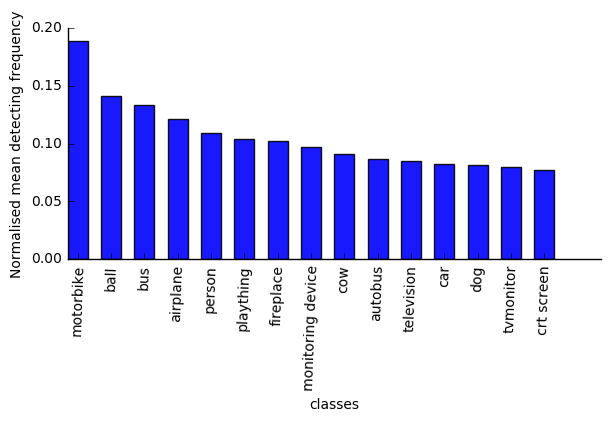}
			\includegraphics[width=0.4\textwidth]{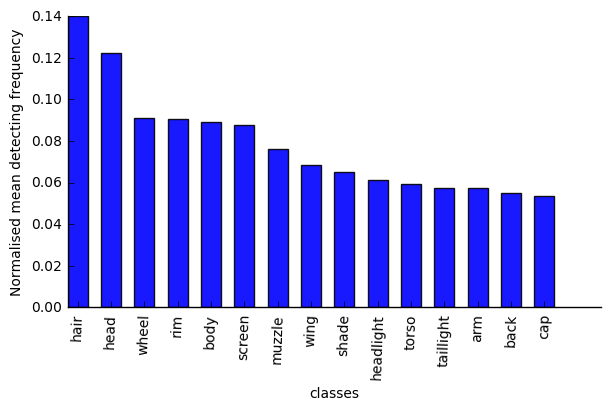}
            \vspace*{-10pt}
		        \caption{Deepnet}
		\end{subfigure}
		\begin{subfigure}{\textwidth}
        \centering
        \includegraphics[width=0.4\textwidth]{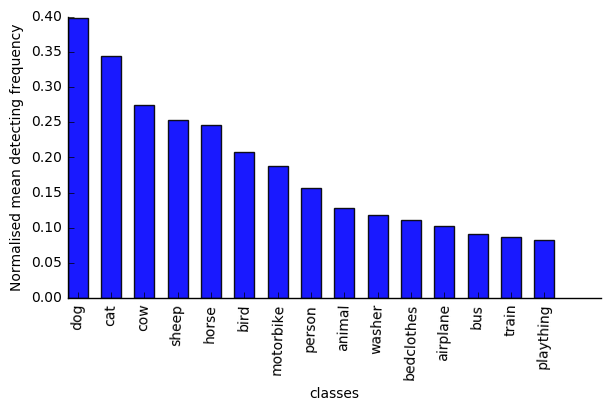}
			\includegraphics[width=0.4\textwidth]{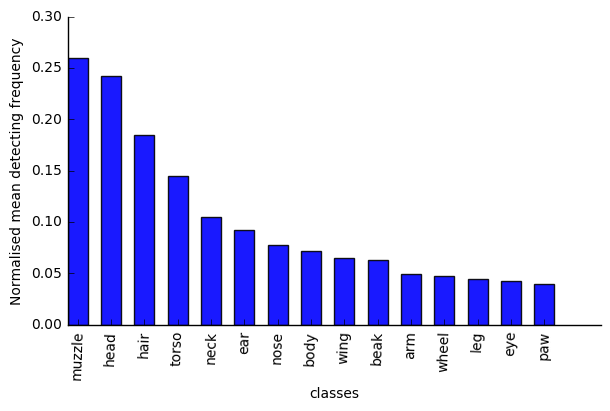}
            \vspace*{-10pt}
			\caption{SalGAN}
		\end{subfigure}
		\begin{subfigure}{\textwidth}
        \centering
        \includegraphics[width=0.4\textwidth]{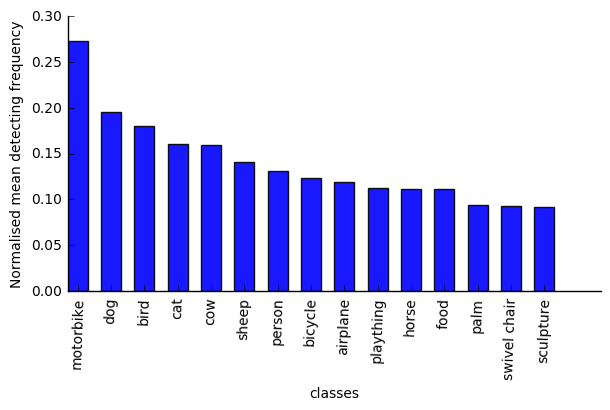}
			\includegraphics[width=0.4\textwidth]{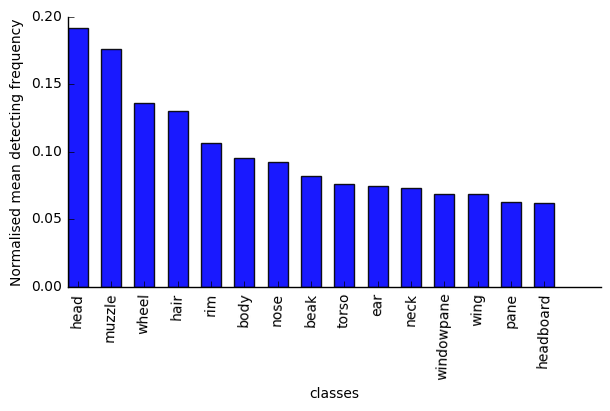}
            \vspace*{-10pt}
			\caption{Saliconc}
		\end{subfigure}
		\begin{subfigure}{\textwidth}
        \centering
			\includegraphics[width=0.4\textwidth]{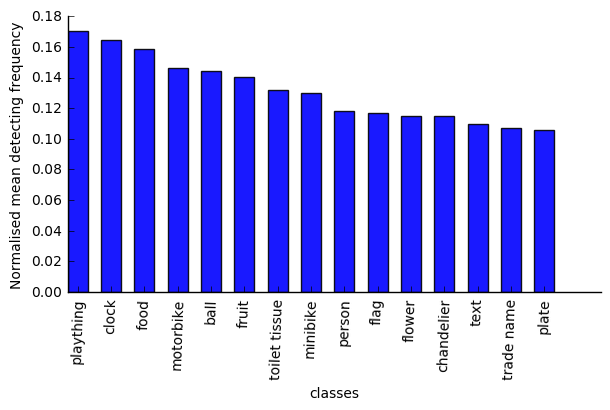}
\includegraphics[width=0.4\textwidth]{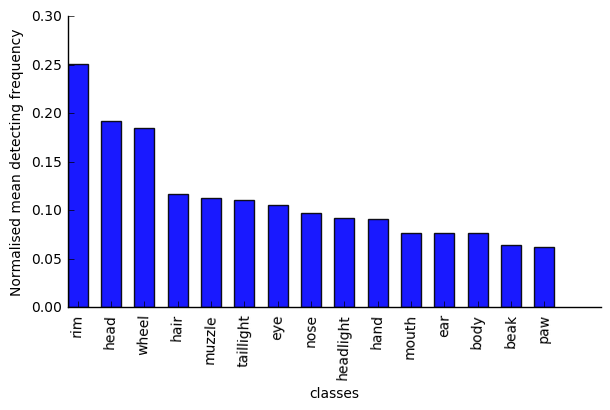}
            \vspace*{-10pt}
			\caption{Saliconf}
		\end{subfigure}
        \begin{subfigure}{\textwidth}
        \centering
            \includegraphics[width=0.4\textwidth]{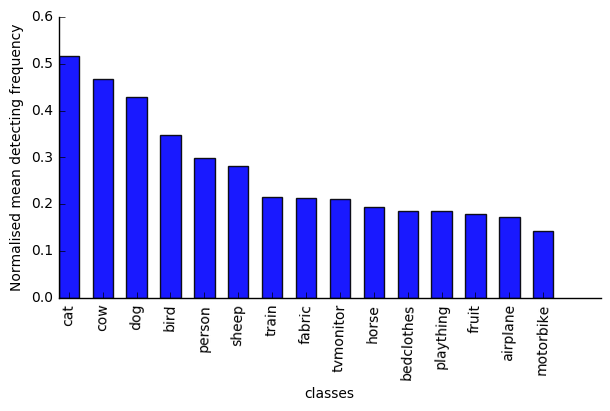}
			\includegraphics[width=0.4\textwidth]{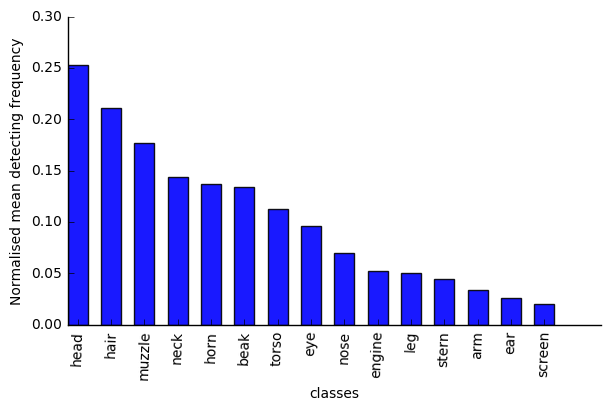}
            \vspace*{-10pt}
			\caption{ours}         
		\end{subfigure}
	\end{center}
    \vspace*{-15pt}
	\caption{The top 15 objects (left) and parts (right) statistics for different saliency models' positive fixation detectors.}
	\label{fig:diss}
\end{figure*}


Figure~\ref{fig:diss} records the normalized mean detection frequency for the analyzed models and for classes (left) and parts (right) in the Broden dataset. 
On the left hand side of the figure shows the results of this analysis for object labels: Most positive fixation detectors are attuned to common animals (dog, cat, cow, sheep, and person). The reason might be that they are fine-tuned from the image recognition models, which have already learnt rich object classes. This assumption is supported by the results on Deepnet, that has been trained from scratch without pre-training, and for which the first four object classes (motorbike, ball, bus, airplane) are not those animals. Interestingly, the detectors on Saliconc and Saliconf are attuned to different visual classes; the coarse model (Saliconc) appears to capture more common object classes than the fine model, as evidenced by higher $f_n$ scores.

            

The right hand side shows similar results but using parts labels instead of object labels. In these graphs, we see that almost all positive fixation detectors focus on the head or head parts (i.e., head, hair, torso, ear, and neck).

\section{Conclusion}
This paper set out to investigate the reason behind the high performance achieved by deep saliency models, compared to shallow models using hand-crafted features based on theoretical considerations about saliency (e.g., center-surround difference).
To this end, we proposed a simple residual-like decoder combined with a pixel-wise exponential absolute distance loss function. The proposed loss function achieves best results among all pixel-wise loss functions and the model performance is on par or better than those state-of-the-art saliency models, despite being based on a simple architecture. Furthermore, we proposed a visualization method for deep gaze prediction models, and did a comprehensive study to reveal the inner representations inside those models. 

Our analyses allow us to draw three conclusions about what is learned by deep saliency models. First, better performing models have developed higher proportions of deep neurons highly predictive of human gaze, and those neurons are attuned to very specific visual patterns.
Second, another category of neurons, which are not predictive of human gaze, appear to encode a form of negative central bias into the model. Third, we have demonstrated that the predictive neurons are attuned to clear semantic categories such as animals (dogs, cats), objects (motorbike, ball) and parts (head, hair). These results provide evidence that the higher prediction performance achieved by deep saliency models is likely caused by the additional semantic content encoded by such networks, allowing the models to capture the fact that specific visual classes are salient in their own right, in contrast to shallow saliency models that rely on low level perceptual patterns (such as center-surround difference). This hints that saliency, as experienced by humans, is a process that likely involves high-level world knowledge in addition to low-level perceptual cues. 

We believe that our results can be useful to measure the gap between current saliency models and the human inter-observer model and to build new models to close this gap. We will share our code to facilitate future research in this direction.

\bibliographystyle{splncs}
\bibliography{egbib}
\end{document}